\documentclass[10pt,journal,compsoc]{IEEEtran}

\usepackage{times}
\usepackage[table,xcdraw]{xcolor} 
\usepackage{booktabs} 
\usepackage{amsmath,amsfonts}
\usepackage{algorithmic}
\usepackage{array}
\usepackage[caption=false,font=normalsize,labelfont=sf,textfont=sf]{subfig}
\usepackage{textcomp}
\usepackage{stfloats}
\usepackage{url}
\usepackage{verbatim}
\usepackage{graphicx}
\usepackage{graphicx} 
\usepackage{amsmath}
\usepackage{amsfonts}
\usepackage{multirow}
\usepackage{color,xcolor}
\usepackage{bbding}
\usepackage{graphicx}
\usepackage{amsmath}
\usepackage{amssymb}
\usepackage{booktabs}
\usepackage{enumitem}
\usepackage{multirow}
\usepackage{setspace} 

\usepackage{algorithm}
\usepackage{algorithmic}

\usepackage[misc]{ifsym} 
\usepackage{newfloat}
\definecolor{lightgray}{gray}{.9}
\definecolor{forestgreen}{RGB}{0,128,69}
\definecolor{cvprpink}{RGB}{219,112,147}
\definecolor{cvpryellow}{RGB}{255, 191, 0}
\definecolor{cvprblue}{rgb}{0.21,0.49,0.74}

\usepackage[colorlinks,
            linkcolor=red,       
            anchorcolor=blue,  
            citecolor=blue,        
            ]{hyperref}

\usepackage{orcidlink}

\usepackage{listings}
\DeclareCaptionStyle{ruled}{labelfont=normalfont,labelsep=colon,strut=off} 
\floatstyle{ruled}
\newfloat{listing}{tb}{lst}{}
\floatname{listing}{Listing}

\usepackage[colorlinks,
            linkcolor=red,       
            anchorcolor=blue,  
            citecolor=blue,        
            ]{hyperref}

\usepackage{amsthm} 

\usepackage{bm}

\usepackage[export]{adjustbox}

\usepackage{threeparttable}
\newcolumntype{I}{!{\vrule width 1pt}}
\makeatletter
\newcommand{\thickhline}{%
    \noalign {\ifnum 0=`}\fi \hrule height 1pt
    \futurelet \reserved@a \@xhline
}
\makeatother
\usepackage{float}
\usepackage{pifont}
\usepackage{ragged2e}

\ifCLASSOPTIONcompsoc
  \usepackage[nocompress]{cite}
\else
  \usepackage{cite}
\fi
\hypersetup{pdftitle={Information Density Imbalance in Visual Object Detection},
pdfauthor={Ziwei Zhao, Yanxi Lu, Yuwei Hu, Shiyang Su, Mingxuan Wang, Chenyue Zhou, Jiayi Chen, Hehan Li, Xiaoshuai Hao, Andi Zhang, Yanbiao Ma},
pdfsubject={Object Detection, Long-Tail Learning, Information Density},
pdfkeywords={Object Detection, Information Density, Long-Tail, Class Bias, Covariance, Fairness}}

\usepackage{enumitem}
\begin{document}
\title{Information Density Imbalance in \\Visual Object Detection}
\author{Ziwei Zhao~\orcidlink{0009-0007-4865-3130},
	  Yanxi Lu,
       Yuwei Hu,
	  Shiyang Su,
        Mingxuan Wang,
        Chenyue Zhou,
	   Jiayi Chen,
        Hehan Li,
        Xiaoshuai Hao,
        Andi Zhang,
        Yanbiao Ma*~\orcidlink{0000-0002-8472-1475}

\thanks{Ziwei Zhao is with the Technical University of Munich. Yanbiao Ma is with the Gaoling School of Artificial Intelligence, Renmin University of China. Chenyue Zhou is with Nanyang Technological University. Hehan Li is with Baidu Inc. Xiaoshuai Hao is with Beijing Academy of Artificial Intelligence. Andi Zhang is with University of Manchester.}
\thanks{Correspondence author: Yanbiao Ma}
\thanks{E-mail: az381@cantab.ac.uk, ybma1998@ruc.edu.cn}
}

\IEEEtitleabstractindextext{%
\begin{abstract}
In object detection, the number of instances is typically used to determine whether a dataset exhibits a long-tailed distribution, implicitly assuming that the model will perform poorly on categories with fewer instances. This assumption has led to extensive research on category bias in datasets with imbalanced instance numbers. However, even in datasets where instance numbers are relatively balanced, models still exhibit category bias, indicating that instance count alone cannot explain this phenomenon. In this work, we first introduce the concept and measurement of information density. We then observe a significant negative correlation between a category’s information density and its accuracy, and we investigate how the training process impacts this relationship. Empirical studies suggest that information density imbalance may be a potential source of category bias. To preliminarily validate the potential of information density, we made simple improvements to three advanced object detection loss functions using this concept. Experiments on the Pascal VOC, COCO-LT, and LVIS datasets demonstrate that information density can significantly reduce model bias while effectively enhancing the overall performance of existing loss functions. This study provides a new perspective for understanding the generalized bias phenomenon in object detection models and offers new tools for designing fairer loss functions and training strategies.
\end{abstract}

}

\maketitle

\IEEEdisplaynontitleabstractindextext

\IEEEpeerreviewmaketitle

\section{Introduction}
\label{sec:intro}

\begin{figure}[t]
\centering
\centerline{\includegraphics[width=1\columnwidth]{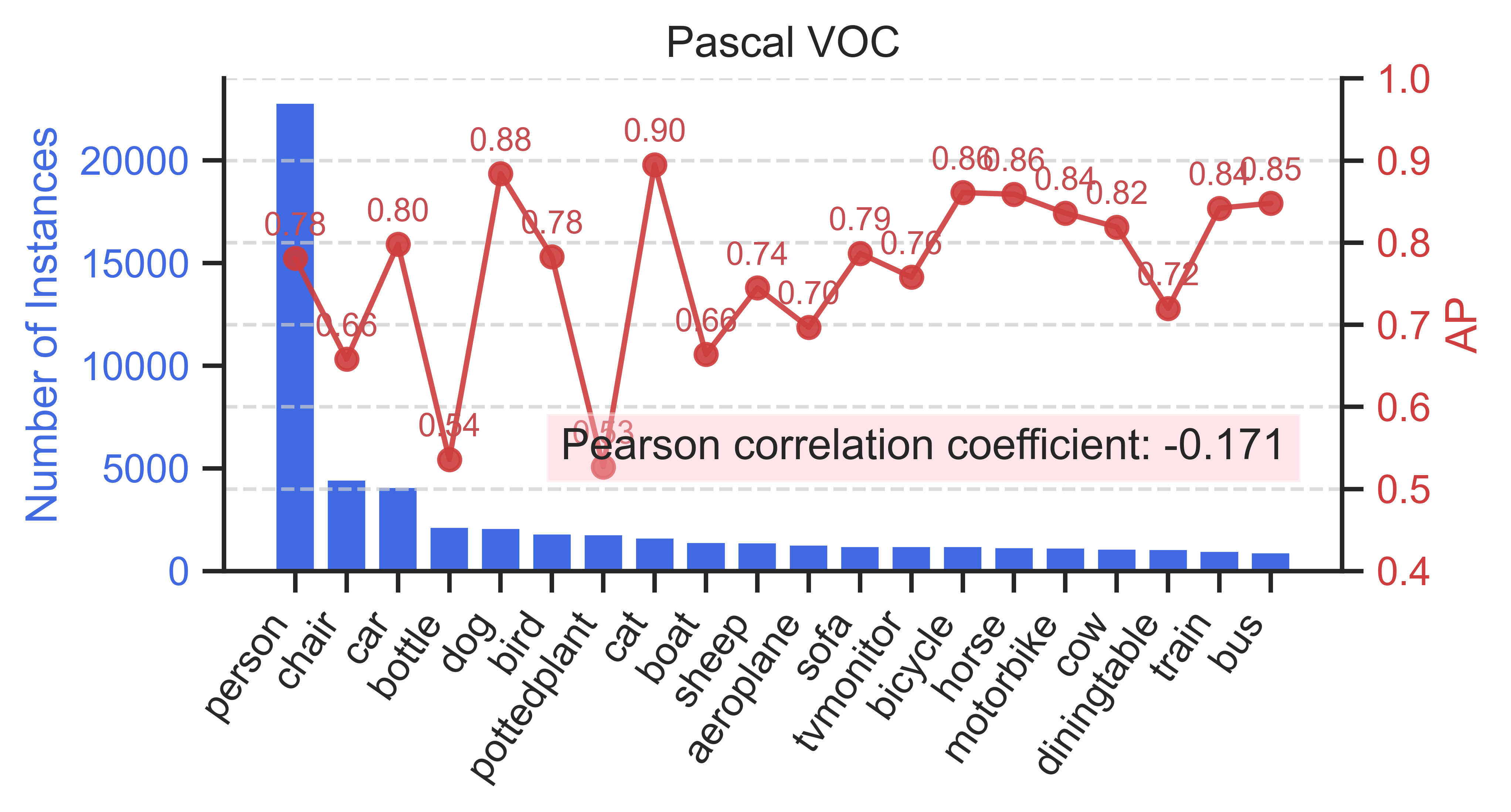}}
\vskip -0.1in
\caption{The left vertical axis represents the number of instances per class. The right vertical axis shows the performance of Faster R-CNN with an R-50-FPN backbone across all classes, trained on the Pascal VOC dataset. The model was trained using the settings described in Section \ref{sec6.2}. The red text box displays the Pearson correlation coefficient between class performance and the number of instances.}
\label{fig1}
\vskip -0.1in
\end{figure}

\IEEEPARstart{O}{bject} detection is a key task in the field of computer vision, widely applied in various practical scenarios such as autonomous driving, medical image analysis, and surveillance systems \cite{jiao2019survey,ijcvsurvey,zou2023object,tkde1}. Despite significant progress in recent years, the issue of dataset imbalance remains a persistent challenge \cite{yang2022survey,tkde4,fur,open,TPAMIimbalance}. Traditional assumptions often attribute the poor performance of models on certain categories to the scarcity of instances, a view that has been widely accepted and has led to many instance-quantity-based improvement methods \cite{number1,tkde3,tkde2,number4,sinha2022class,lvis,number5}. However, our quantitative research suggests that this assumption may be overly simplistic. As shown in Fig.~\ref{fig1}, empirical analysis on the Pascal VOC dataset \cite{VOC} indicates that the Pearson correlation coefficient between the number of category instances and corresponding detection accuracy is only -0.171. Such a weak correlation prompts us to hypothesize that other deeper factors may be influencing the performance of object detection models.

In the field of image classification, multiple studies \cite{sinha2022class,ma2023delving,ma2023curvature,icml2024balanced,ma2024unveiling} have observed that classification models still exhibit significant bias towards certain categories even when the sample quantities are perfectly balanced. To uncover the underlying mechanisms, \cite{ma2023delving,ma2023curvature,ma2024unveiling} have found a significant correlation between the geometric properties of perceptual manifolds in deep neural networks and model bias. \cite{icml2024balanced} discovered that the decay rate of the feature spectrum of the embedding distribution could be a source of model bias. However, in the task of object detection, category bias induced by non-long-tailed distributions has been relatively overlooked. Therefore, we aim to explore and quantify the deeper factors within datasets that influence model bias, beyond merely the number of instances.

Intuitively, if a category has highly diverse instances, it becomes more difficult for the model to learn this category. Conversely, if a category's diversity is low, the model can learn it well with fewer instances. Additionally, a more intuitive factor is the size of the instances: when a category contains instances of larger size, the model can detect them more easily \cite{size2}. However, quantifying instance diversity is the first challenge in verifying this intuition. Inspired by the neural manifolds generated during visual processing in the human brain \cite{manifold1,manifold2,manifold3}, we propose methods for measuring instance diversity and calculating the total area of instances in Section \ref{sec4.1}. Our findings indicate that both instance diversity and the total area of instances are highly correlated with category accuracy, validating our hypothesis.

Furthermore, we introduce the concept of information density, defined as the ratio of instance diversity to the total area of instances (Section \ref{sec4.2}). Information density represents the amount of information per unit area, reflecting the concentration of information. Surprisingly, there is a significant negative correlation between information density and category detection accuracy (see Fig.~\ref{fig2} and Fig.~\ref{fig3}). Moreover, the correlation between information density and category accuracy strengthens throughout training (see Fig.~\ref{fig4}), indicating that current optimization objectives overlook the model bias induced by imbalances in information density. Information density effectively quantifies the combined impact of category diversity and instance area, providing a new perspective for understanding the generalized bias phenomenon in object detection models. This concept also offers a theoretical foundation for designing more equitable loss functions and training strategies.

To demonstrate the potential of information density, we improved three advanced loss functions for object detection in Section \ref{sec5.1}: Seesaw Loss \cite{seesaw}, Equalized Focal Loss \cite{efl}, and C2AM Loss \cite{c2am}. Additionally, to dynamically update information density during training, we proposed a novel low-cost training strategy in Section \ref{sec5.2}. Comprehensive evaluations on non-long-tailed (Pascal VOC \cite{VOC}) and long-tailed (COCO-LT \cite{number4}, LVIS v1.0 \cite{lvis}) datasets show that incorporating information density into the loss function significantly alleviates model bias and improves overall performance.
In summary, our research emphasizes the importance of understanding the factors influencing object detection performance beyond instance quantity. Future research can further explore how to optimize model training based on information density to enhance model performance in handling categories with high diversity and small instance areas.

\section{Related Work}

In recent years, significant progress has been made in object detection technology. Mainstream research has focused on developing general object detection models, while a few studies have been proposed to improve overall performance by mitigating bias in object detection models \cite{seesaw,efl,c2am,eql,eqlv2,ecm}, with Focal Loss \cite{focal} and Seesaw Loss \cite{seesaw} being two typical examples. To specifically address the issue of model bias, researchers have introduced the long-tailed object detection task. This task has both pros and cons; while it artificially constructs long-tailed object detection datasets to amplify model bias, providing a challenging scenario \cite{number4,lvis}, it simultaneously leads researchers to overlook the bias exhibited by models in more balanced object detection scenarios. This widely observed model bias cannot be fully explained by the imbalance in instance numbers, and there is a lack of research on this phenomenon in the field of object detection. 

In the following, we will first introduce the current state and progress of research in the field of long-tailed object detection, and then discuss existing methods for measuring the difficulty of class learning. In our discussion of class difficulty measurement, we will comprehensively review relevant advancements, not limited to object detection tasks.

\subsection{Long-Tailed Object Detection}
\label{sec2.1}

In the research of long-tailed object recognition, the main approaches include data re-sampling \cite{tkde6,reb7}, specialized loss function design \cite{number3,sinha2022class}, architectural improvements \cite{RIDE,tkde7}, decoupled training \cite{BBN,decoupling}, and data augmentation \cite{tkde5,CMO}.

Data re-sampling is a common method to address imbalanced datasets by increasing the sampling frequency of tail class samples to balance the data distribution \cite{tkde6}. Common re-sampling strategies include Class-aware sampling \cite{Class_aware_sampling} and Repeat factor sampling (RFS) \cite{lvis}. These methods can be employed at different stages of training to achieve a multi-stage training process. Specialized loss function design is another technical approach to tackling long-tailed challenges. For instance, EQL \cite{eql} reduces suppression on tail classes by truncating the negative gradients from head classes. The subsequent EQLv2 \cite{eqlv2} further improves this approach through a gradient balancing mechanism. Other methods, such as Seesaw Loss \cite{seesaw}, Equalized Focal Loss \cite{efl}, ACSL \cite{acsl}, and LOCE \cite{loce}, reduce excessive suppression of tail classes by dynamically adjusting classification logits or suppressing overconfident scores. C2AM \cite{c2am} observed that the severe imbalance in weight norms across classes leads to pathological decision boundaries, and therefore proposes learning fairer decision boundaries by adjusting the ratio of weight norms. 

Current research mainly focuses on these two directions. In addition, module improvement emphasizes modifying the structure of detectors to address long-tailed distribution issues. For example, BAGS \cite{bags} and Forest R-CNN \cite{forest} mitigate the impact of head classes on tail classes by grouping all classes based on valuable prior knowledge. Decoupled training \cite{decoupling} has found that long-tailed distributions do not significantly affect the learning of high-quality features, thus some methods freeze the feature extractor parameters during the classifier learning phase, adjusting only the classifier \cite{fdc,number4,zhang2021distribution}. Data augmentation, as a means of introducing additional sample variability, has been shown to provide further improvements in long-tailed detection tasks. Recently proposed methods such as Simple Copy-Paste \cite{copy_paste}, FDC \cite{fdc}, FASA \cite{fasa}, and FUR \cite{fur} supplement the insufficiency of tail-class samples by performing data augmentation in both image and feature spaces \cite{tkde5}.

\subsection{Methods for Measuring Class Difficulty}
\label{sec2.2}

\textbf{The study of class difficulty is most relevant to our work.} Most research addressing class bias has focused on scenarios with sample imbalance, where rebalancing strategies based on sample size can be somewhat effective. However, recent studies have reported that even when sample sizes are perfectly balanced, classification models still exhibit significant performance disparities across different classes. Investigating the root causes of model bias in scenarios where sample sizes are balanced is crucial for improving model fairness and understanding learning mechanisms. However, research on this issue is still limited. From a geometric perspective, DSB \cite{ma2023delving}, CR \cite{ma2023curvature}, and IDR \cite{ma2024unveiling} conceptualize the data classification process as the disentangling and separating of different perceptual manifolds. These three studies respectively reveal that the geometric properties of perceptual manifolds—volume, curvature, and intrinsic dimensionality—are significantly correlated with class performance. \cite{icml2024balanced} discovered that differences in the spectral features of classes could be a source of class bias. Unfortunately, in the field of object detection, there has been no research exploring the underlying causes of model bias.
Our work is the first to directly report on the widespread bias present in object detection models and to attempt to explore the potential mechanisms underlying this bias.

\section{Information Density Imbalance}
\label{sec4}

In Section \ref{sec4.1}, we first introduce how to calculate the total area of instances in a category, and then propose a measurement to estimate the diversity of instances in a category. In Section \ref{sec4.2}, we present two findings: the significant negative correlation between instance diversity and category average precision, and the significant positive correlation between the total area of instances and category accuracy. Based on these observations, we propose the concept of information density to unify these two factors. Finally, in Section \ref{sec4.3}, we explore the impact of learning on the correlation between information density and category average precision.

\subsection{Measurement of Instance Diversity and Total Area}
\label{sec4.1}

\subsubsection*{Definition and Measurement of Total Area of Instances}
\label{sec4.1.1}

The total area of instances in a category can be obtained by summing the ground truth areas of all instances. Given a set of ground truth bounding boxes $B=\{B_1, B_2,\dots,B_m\}$ for all instances in a category, where $B_i$ represents the ground truth bounding box of instance $i$, the area $A(B_i)$ of each bounding box can be calculated using the following formula:
\begin{equation}
\begin{split}
A(B_i) = (x_{max}-x_{min})\cdot (y_{max}-y_{min}),
\nonumber
\end{split}
\end{equation}
where $(x_{max}, y_{max})$ and $(x_{min}, y_{min})$ are the coordinates of the top-right and bottom-left corners of the bounding box $B_i$, respectively. The total area of the category can be obtained by summing the areas of all instances:
\begin{equation}
\begin{split}
A(B) = \sum_{i=1}^{m}A(B_i). 
\nonumber
\end{split}
\end{equation}

Next, we introduce how to measure the diversity of instances within a category.

\subsubsection*{Definition and Measurement of Instance Diversity}
\label{sec4.1.2}

In the nervous system, when neurons are stimulated by physical features of the same category, a perceptual manifold is formed \cite{manifold2,manifold4}. The formation of the perceptual manifold helps the nervous system differentiate and process objects with different features within the same category. Recent studies \cite{manifold5,ma2023delving,ma2024unveiling} have shown that the response of deep neural networks to images is similar to human vision, following the manifold distribution law, meaning that the embeddings of a category of natural images are distributed near a low-dimensional perceptual manifold embedded in a high-dimensional space. Continuous sampling along one dimension of the manifold corresponds to continuous changes in physical features. Therefore, the volume of the perceptual manifold mapped by deep neural networks can measure the diversity of a category.

Given a category, suppose the embeddings of all instances are \(X=[x_1,\dots,x_m] \in \mathbb{R}^{p \times m}\), where \(p\) is the dimension of the embeddings, and \(m\) is the number of instances. It is important to note that the embeddings used to calculate diversity need to be extracted from the classification module of the object detection model, not the regression module. The volume of the perceptual manifold corresponding to the embedding set \(X\) can be measured by the span of the subspace spanned by the eigenvectors of the covariance matrix of \(X\). However, directly calculating the sample covariance matrix in high-dimensional situations is often inaccurate, so we use the Ledoit-Péché nonlinear shrinkage method \cite{shrinkage_method} to improve the estimation results.

First, the sample covariance matrix is estimated as:
\begin{equation}
\begin{split}
\Sigma(X)=\frac{1}{m}(X-\bar{X})(X-\bar{X})^T,
\nonumber
\end{split}
\end{equation}
where $\bar{X}$ is the mean vector of the data:
\begin{equation}
\begin{split}
\bar{X}=\frac{1}{m}\sum_{i=1}^{m}x_i. 
\nonumber
\end{split}
\end{equation}
Next, the sample covariance matrix $\Sigma(X)$ is decomposed into eigenvalues $\lambda_i,i=1,\dots,p$ and the corresponding eigenvector matrix $V$:
\begin{equation}
\begin{split}
\Sigma(X)=V\Lambda V^T,
\nonumber
\end{split}
\end{equation}
where $\Lambda$ is a diagonal matrix with the eigenvalues $\lambda_i,i=1,\dots,p$ on the diagonal.

According to the Marčenko-Pastur distribution in random matrix theory \cite{donoho2018optimal}, a nonlinear transformation is applied to the eigenvalues. Specifically, the parameter $c=p/m$ is calculated of the Marčenko-Pastur distribution. According to the Marčenko-Pastur distribution, the theoretical support interval of the eigenvalues is
\begin{equation}
\begin{split}
\lambda_{\pm}=(1\pm \sqrt{c})^2.
\nonumber
\end{split}
\end{equation}
Then, the following nonlinear transformation is applied to each eigenvalue:
\begin{equation}
\begin{split}
d_i=\max(\lambda_i, \lambda_-), i=1,\dots,p.
\nonumber
\end{split}
\end{equation}
The purpose of this nonlinear transformation is to avoid the undue influence of excessively small eigenvalues on the covariance matrix estimation, ensuring that all eigenvalues are at least above the theoretical lower bound, thus improving the stability of the estimation. Finally, the covariance matrix is reconstructed using the nonlinearly transformed eigenvalues $d_i$ and the eigenvector matrix $V$:
\begin{equation}
\begin{split}
\Sigma_{LP}=Vdiag(d_i, i=1,\dots,p)V^T.
\nonumber
\end{split}
\end{equation}
Furthermore, we formally define the instance diversity of a category as
\begin{equation}
\begin{split}
Vol(X)=\frac{1}{2}\log_2\det(I+\Sigma_{LP}),
\nonumber
\end{split}
\end{equation}
where $\det(I+\Sigma_{LP})$ is the determinant of the matrix $I+\Sigma_{LP}$. The determinant, equivalent to the product of its eigenvalues, represents the span of the subspace spanned by the eigenvectors. The use of the logarithmic transformation is to enhance the stability and smoothness of the computation results. This method of measuring instance diversity provides a tool for the quantitative study of the generalized imbalance phenomenon in object detection models.

\begin{algorithm}[t]
\setstretch{1.05} 
\caption{Information Density Calculation}
\label{alg:info_density}
\begin{algorithmic}[1]
\STATE \textbf{Input:} Set of bounding boxes $B = \{B_1, B_2, \dots, B_m\}$, feature embeddings $X = [x_1, x_2, \dots, x_m] \in \mathbb{R}^{p \times m}$

\STATE \textbf{Compute Total Area $A(B)$:}
\FOR{$i = 1$ to $m$}
    \STATE $A(B_i) \gets (x_{\max}^i - x_{\min}^i) \times (y_{\max}^i - y_{\min}^i)$
\ENDFOR
\STATE $A(B) \gets \sum_{i=1}^{m} A(B_i)$

\STATE \textbf{Compute Diversity $Vol(X)$:}
\STATE Compute sample covariance matrix: $\Sigma(X) \gets \frac{1}{m} (X - \bar{X})(X - \bar{X})^T$, where $\bar{X} = \frac{1}{m} \sum_{i=1}^{m} x_i$
\STATE Perform eigenvalue decomposition on $\Sigma(X)$: \\$\Sigma(X) = V \Lambda V^T$
\STATE Compute Marčenko-Pastur distribution parameter: \\$c \gets \frac{p}{m}$
\STATE Compute theoretical support interval: $\lambda_{\pm} \gets (1 \pm \sqrt{c})^2$
\FOR{$i = 1$ to $p$}
    \STATE $d_i \gets \max(\lambda_i, \lambda_{-})$
\ENDFOR
\STATE Reconstruct covariance matrix with transformed eigenvalues: $\Sigma_{\text{LP}} \gets V \, \text{diag}(d_i) \, V^T$
\STATE Compute diversity: $Vol(X) \gets \frac{1}{2} \log_2 \text{det}(I + \Sigma_{\text{LP}})$

\STATE \textbf{Compute Information Density $I$:}
\STATE $I \gets \frac{Vol(X)}{A(B)}$

\STATE \textbf{Output:} $I$
\end{algorithmic}
\end{algorithm}
\vspace{-20pt}

\begin{figure*}[t]
\centering
	\begin{minipage}{0.495\linewidth}
		\centering
		\includegraphics[width=1\linewidth]{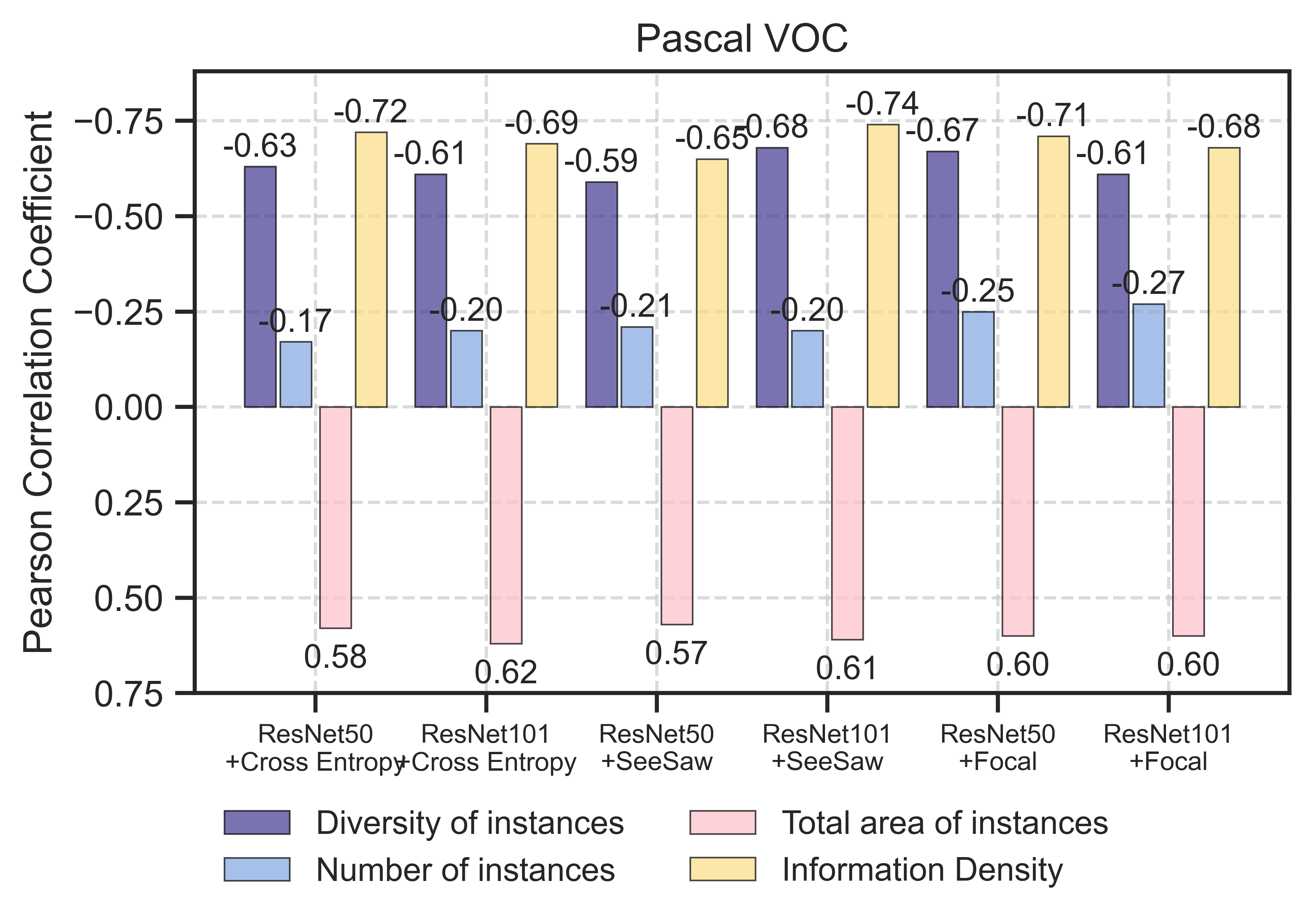}
	\end{minipage}
	\begin{minipage}{0.495\linewidth}
		\centering
		\includegraphics[width=1\linewidth]{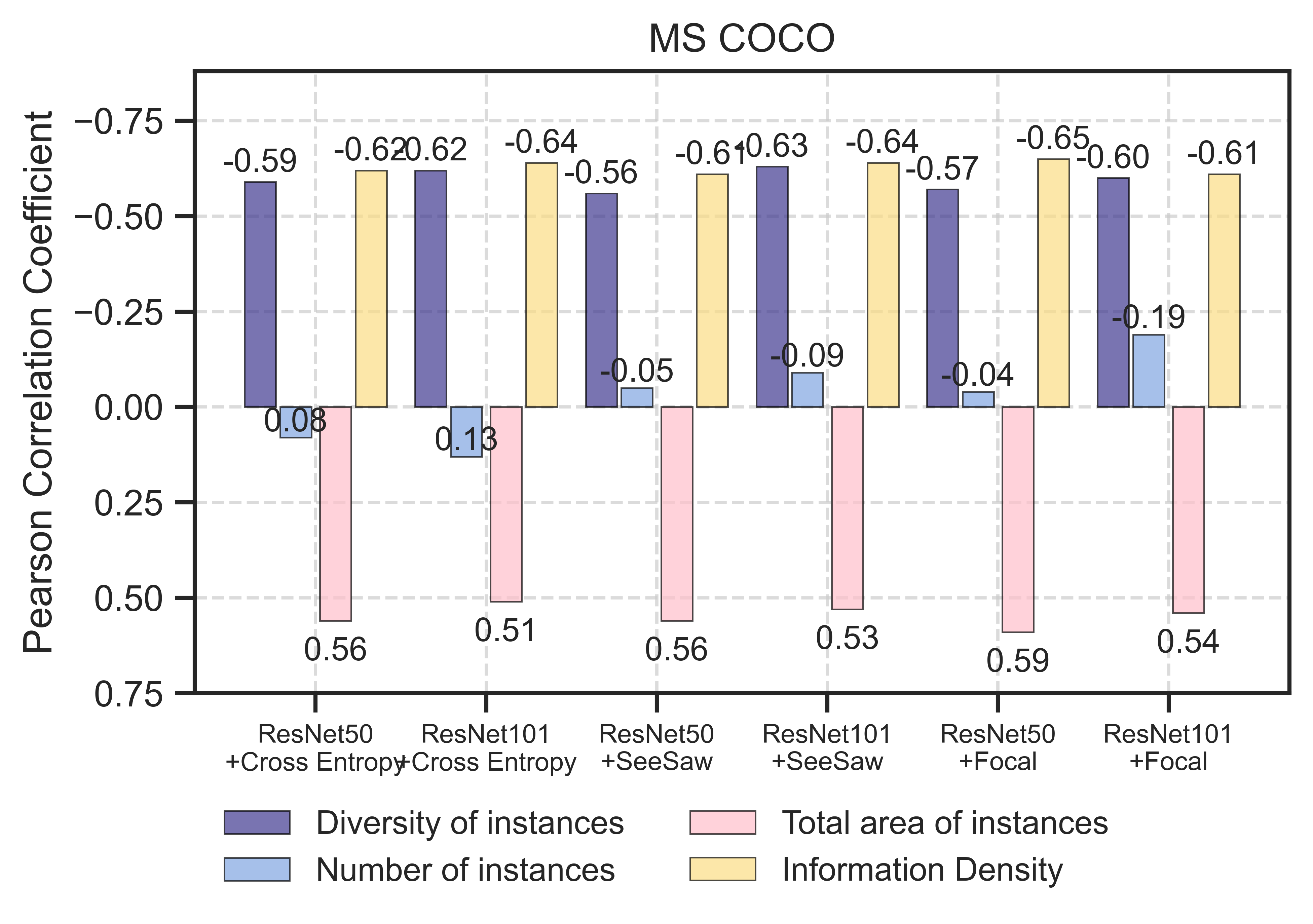}
	\end{minipage}
\vskip -0.1in
\caption{The Pearson correlation coefficients between the average precision (AP) per category on the Pascal VOC dataset and the following factors: instance diversity, total instance area, instance count, and information density. Note that all four factors are measured at the category level.}
\label{fig2}
\vskip -0.1in
\end{figure*}

\vspace{6pt}
\subsection{Imbalance of Information Density and Model Bias}
\label{sec4.2}

We trained standard Faster R-CNN \cite{faster} models with R-50-FPN and R-101-FPN \cite{resnet,size2} backbones on Pascal VOC \cite{VOC} and MS COCO \cite{MSCOCO}, using cross-entropy (CE), SeeSaw, and Focal loss functions. We used each instance's ground truth as the Region of Interest (ROI) and extracted classification features. We then computed instance diversity for each category and analyzed the Pearson correlation between instance diversity and category average precision. The experimental results, as shown in Fig.~\ref{fig2}, indicate a significant negative correlation between category diversity and accuracy, suggesting that the model's learning difficulty increases with higher intra-category variability.
Conversely, we found a significant positive correlation between the total area of instances in a category and detection accuracy. Categories with larger instance areas are more accurately detected by the model. These results indicate that intra-category variability and instance area significantly impact the model's learning performance.

\begin{figure}[t]
\centering
\centerline{\includegraphics[width=1\columnwidth]{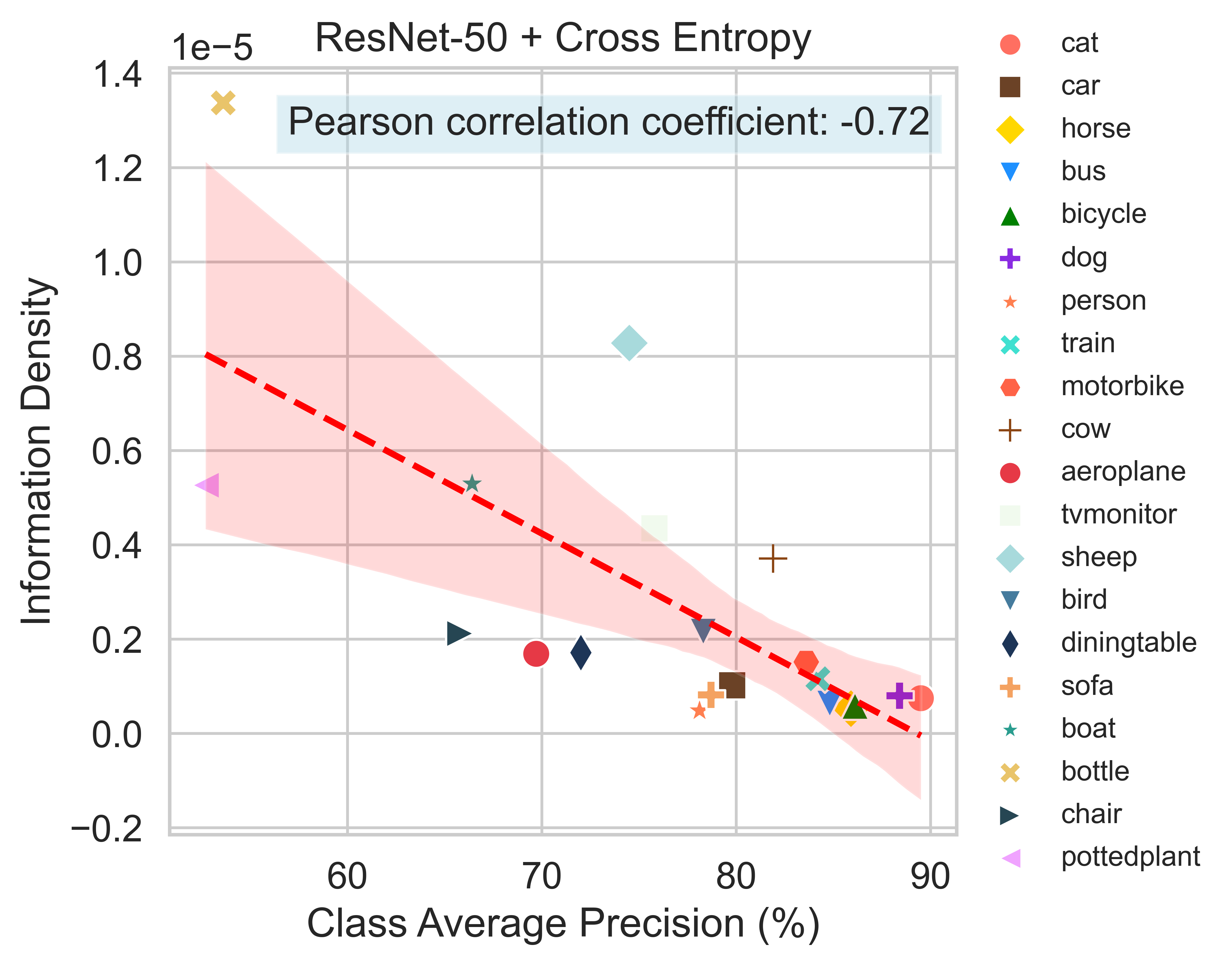}}
\vskip -0.1in
\caption{Taking Faster R-CNN trained with CE loss as an example, the horizontal axis represents the average precision of each category, and the vertical axis represents the information density of each category. The blue text box reports the Pearson correlation coefficient between the average precision of each category and its information density.}
\label{fig3}
\vskip -0.1in
\end{figure}

Further, we combined these two factors, instance diversity and total area, to define information density:
\begin{equation}
\begin{split}
I=\frac{Vol(X)}{A(B)},
\nonumber
\end{split}
\end{equation}
where $X=[x_1,x_2,\dots,x_m]\in \mathbb{R}^{p\times m}$ represents the set of embeddings for all instances in a category, and $Vol(X)$ denotes instance diversity. $A(B)$ represents the total area of all ground truth bounding boxes. Clearly, information density $I$ represents the amount of information per unit area, reflecting the richness of information. Higher information density implies that the model needs to process more diverse features within a relatively small region, increasing the complexity and difficulty of learning.

We were surprised to find that information density $I$ exhibits an even more significant negative correlation with category average precision (see Fig.~\ref{fig2} and Fig.~\ref{fig3}), and this metric has a physical meaning. Information density $I$ effectively quantifies the combined impact of category diversity and instance area, providing a unified measurement standard.

This empirical study suggests directions for precisely mitigating model bias. For example, given two categories with the same number of instances but different levels of diversity, the category with higher diversity should obviously receive more attention. However, traditional re-weighting methods based on instance count fail to achieve this. In contrast, the information density can more accurately guide the model to handle different categories more fairly.


\subsection{Training Increases the Correlation Between Category Information Density and Average Precision}
\label{sec4.3}

\begin{figure*}[t]
\centering
	\begin{minipage}{0.497\linewidth}
		\centering
		\includegraphics[width=1\linewidth]{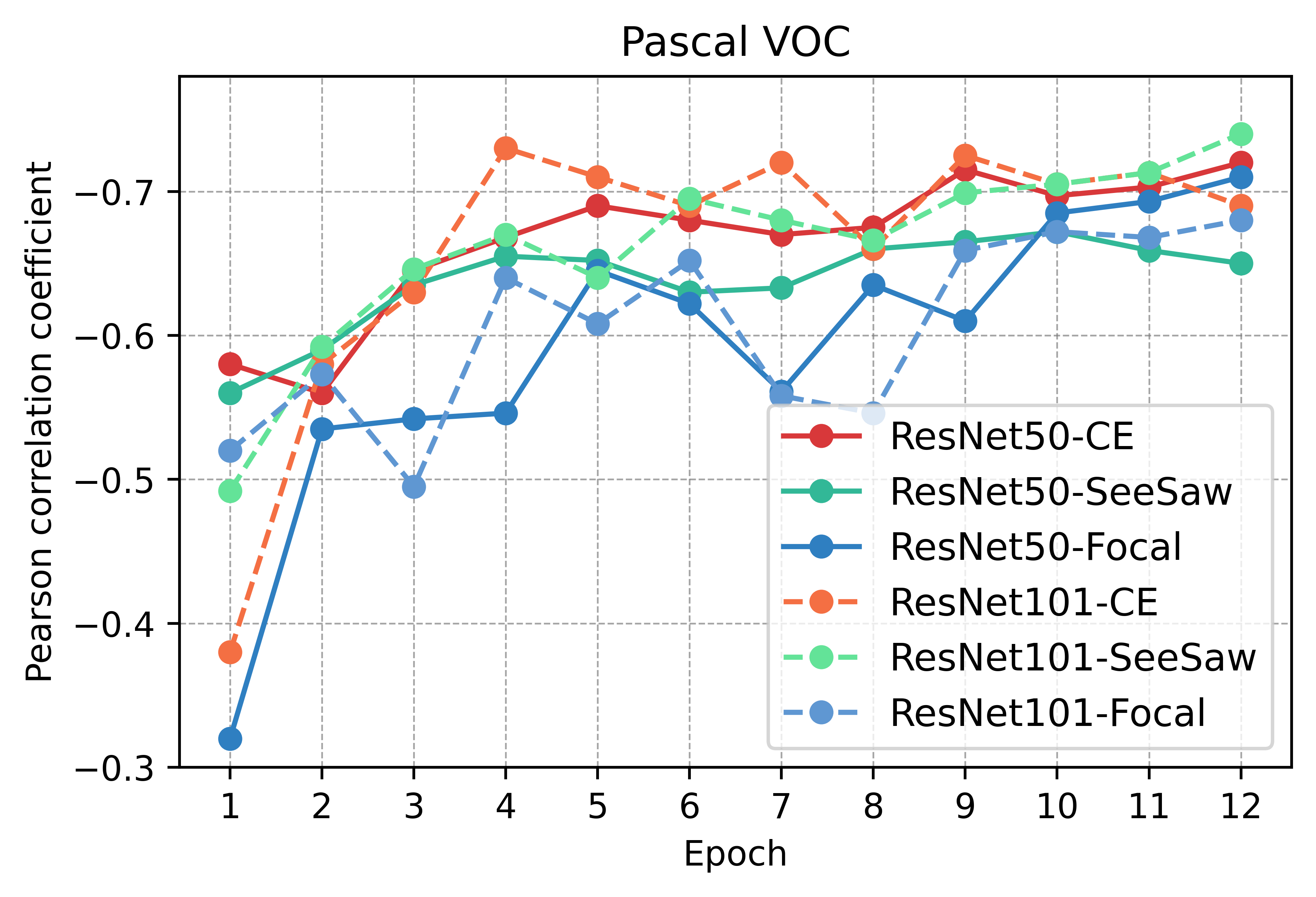}
	\end{minipage}
	\begin{minipage}{0.497\linewidth}
		\centering
		\includegraphics[width=1\linewidth]{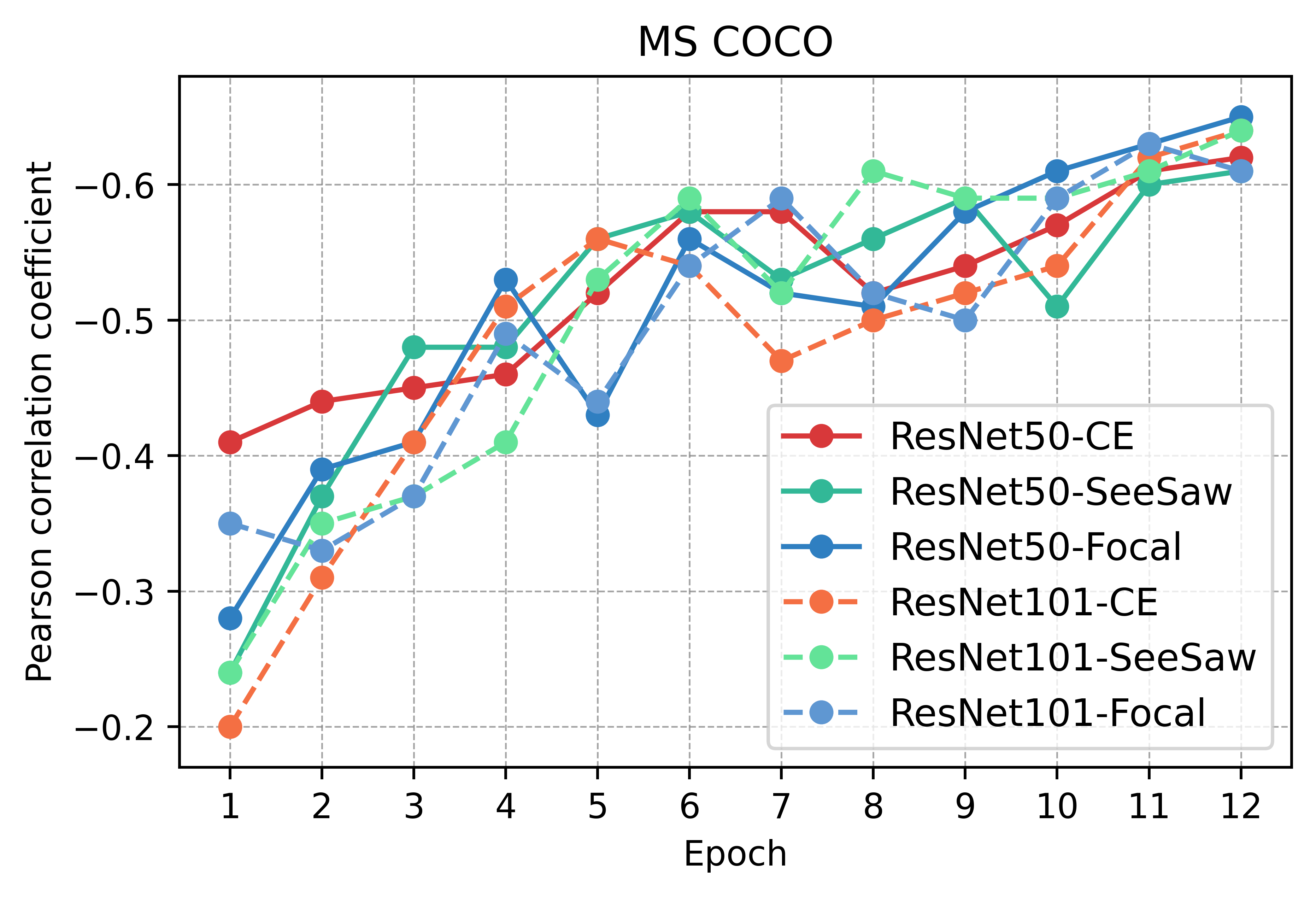}
	\end{minipage}
\vskip -0.15in
\caption{The Pearson correlation coefficient between category information density and average precision (AP) per category over epochs.}
\label{fig4}
\vskip -0.1in
\end{figure*}

Interestingly, our experiments also reveal that the correlation between information density and category average precision gradually strengthens as training progresses (as shown in Fig.~\ref{fig4}). This phenomenon may be attributed to the model's tendency to initially learn the more prominent and common features within the dataset during the early stages of training. At this stage, the model has yet to fully understand and distinguish complex features, resulting in a weaker influence of information density on class accuracy. As training continues, the model progressively learns and extracts more intricate and nuanced features. For categories with higher information density, the model needs to identify and differentiate a greater number of distinct features within a given area, which increases the difficulty of learning. 

This trend persists until the end of training, where the significant correlation between information density and category accuracy remains. This indicates that the existing optimization objectives insufficiently address categories with high information density. To alleviate this issue and improve the overall performance of the model, we will introduce targeted improvements to the existing object detection loss functions in the next section. These improvements aim to better balance the learning difficulty across different categories, particularly by giving more attention to categories with high information density. Through this approach, we hope to reduce model bias and achieve better performance in handling complex and highly diverse categories.

\section{Method}
\label{sec5}

In this section, we first demonstrate how to embed information density into three advanced object detection loss functions (SeeSaw, EFL, and C2AM Loss). Then, we propose a novel training strategy to achieve more efficient dynamic updates of information density during the learning process.

\subsection{Embedding Information Density into Loss Function}
\label{sec5.1}

\subsubsection{Integration of Information Density into SeeSaw Loss}
\label{sec5.1.1}

The original purpose of SeeSaw Loss \cite{seesaw} is to reduce the negative gradient exerted by head class samples on tail classes, thereby increasing the model's attention to tail classes. The form of Seesaw loss is:
\begin{equation}
\begin{split}
L_{seesaw}(z)=-\sum_{i=1}^{C}y_i\log(\sigma_i), 
\nonumber
\end{split}
\end{equation}
with 
\begin{equation}
\begin{split}
\sigma_i=\frac{e^{z_i}}{\sum_{j\ne i}^{C} S_{ij}e^{z_j}+e^{z_i}}.
\nonumber
\end{split}
\end{equation}
Here, \( z = [z_1, z_2, \ldots, z_C] \) and \( \sigma = [\sigma_1, \sigma_2, \ldots, \sigma_C] \) are the predicted logits and probabilities of the classifier, respectively. \( y_i \) is the true label of the sample. \( S_{ij} \) is the balancing factor between different classes, composed of the mitigation factor and the compensation factor, specifically \( S_{ij} = M_{ij} \cdot C_{ij} \). The mitigation factor \( M_{ij} \) is defined as the ratio of the number of instances in class \( j \) to class \( i \). When the number of instances in class \( i \) is greater than that in class \( j \), the negative gradient of class \( j \) is suppressed. The compensation factor focuses on misclassified samples; when a sample from class \( i \) is incorrectly identified as class \( j \), the negative gradient generated for class \( j \) increases.

Since we observed that the model performs poorly on classes with high information density, we want the model to pay more attention to these classes. Assuming the information densities of \( C \) classes are \( I_1, I_2, \ldots, I_C \), we normalize the information density to the \([0, 1]\) range as follows:
\begin{equation}
\begin{split}
I_i = \frac{e^{I_i /(\bar{I}\cdot \sqrt{C} )}}{\sum_{j=1}^{C} e^{I_j / (\bar{I}\cdot \sqrt{C})}}\cdot C+1, \quad \bar{I} = \sum_{i=1}^{C} I_i.
\nonumber
\end{split}
\end{equation}

Furthermore, we propose modifying the balancing factor \( S_{ij} \) to:
\begin{equation}
\begin{split}
S_{ij} = \left( \frac{I_i}{I_j} \right)^k \cdot M_{ij} \cdot C_{ij},
\nonumber
\end{split}
\end{equation}
where \( I_i \) represents the information density of class \( i \), and \( k \) is a hyperparameter. The gradient of \( L_{\text{seesaw}}(z) \) with respect to the negative class \( j \) on \( z_j \) is:
\begin{equation}
\begin{split}
\frac{\partial L_{\text{seesaw}}(z)}{\partial z_j} = \left( \frac{I_i}{I_j} \right)^k \cdot M_{ij} \cdot C_{ij} \cdot \frac{e^{z_j}}{e^{z_i}} \cdot \sigma_i.
\nonumber
\end{split}
\end{equation}

When \( I_i \) is greater than \( I_j \), the model needs to give more attention to class \( i \), thus increasing the negative gradient for class \( j \); conversely, it suppresses the negative gradient for class \( j \). Other details remain consistent with the original Seesaw loss. We refer to the modified loss function as Information Density Guided Seesaw Loss (IDG-Seesaw).

\subsubsection{Information Density Guided Focal Loss (IDG-FL)}
\label{sec5.1.2}

Since Focal Loss cannot differentially handle different classes, Equalized Focal Loss (EFL) \cite{efl} proposes a class-specific focusing factor to address the imbalance of positive and negative samples across different classes. The EFL loss function is defined as:
\begin{equation}
\begin{split}
EFL(p_t) = -\sum_{i=1}^{C} \alpha_t \left( \frac{\gamma_b + \gamma_v^i}{\gamma_b} \right) (1 - p_t)^{\gamma_b + \gamma_v^i} \log(p_t),
\nonumber
\end{split}
\end{equation}
where the meanings of \( \alpha_t \) and \( p_t \) are consistent with those in focal loss. The focusing factor \( \gamma^i \) consists of class-independent and class-dependent parameters: \( \gamma^i = \gamma_b + \gamma_v^i = \gamma_b + s(1 - g^i) \). Increasing the focusing factor reduces the loss, so the weighting term \( \left( \frac{\gamma_b + \gamma_v^i}{\gamma_b} \right) \) in EFL is used to enhance the absolute magnitude of the loss for each class. However, EFL still does not consider the model bias caused by the imbalance of information density between classes. To better address this issue, we propose an improved method that incorporates information density into EFL. Specifically, we modify the weighting term \( \left( \frac{\gamma_b + \gamma_v^i}{\gamma_b} \right) \) to:
\begin{equation}
\begin{split}
\left( \frac{\gamma_b + \gamma_v^i}{\gamma_b} \right) \cdot I_i,
\nonumber
\end{split}
\end{equation}
where
\begin{equation}
\begin{split}
I_i = \frac{e^{I_i /(\bar{I}\cdot \sqrt{C} )}}{\sum_{j=1}^{C} e^{I_j / (\bar{I}\cdot \sqrt{C})}}\cdot C+1, \quad \bar{I} = \sum_{i=1}^{C} I_i.
\nonumber
\end{split}
\end{equation}

By introducing information density \( I_i \), we can more effectively account for the imbalance of information density between classes, thereby reducing model bias. Apart from this modification, other details remain consistent with the original Equalized Focal Loss (EFL).

\subsubsection{Information Density Guided C2AM Loss (IDG-C2AM)}
\label{sec5.1.3}

In long-tailed object detection tasks, Category-Aware Angular Margin Loss (C2AM) \cite{c2am} discovered that the norm distribution of classifier weights is highly imbalanced, leading to severe compression of the decision space for classes with smaller weight norms. To address this issue, C2AM proposed that classes with greater diversity should occupy larger decision spaces. However, due to the inability to directly quantify class diversity, C2AM chose to smooth the ratios between class weight norms instead of the diversity ratios and used these ratios to adjust the optimization objective, thus reasonably partitioning the decision space. Specifically, the C2AM loss function is formulated as follows:
\begin{equation}
\begin{split}
L_{C2AM}=-\log(\frac{e^{s\cdot\cos(\theta_i)}}{e^{s\cdot\cos(\theta_i)+\sum_{j=1,j\ne i}^{C}e^{s\cdot\cos(\theta_j+m_{ij})}}}),
\nonumber
\end{split}
\end{equation}
where
\begin{equation}
\begin{split}
m_{ij}=\max(0,\frac{\alpha}{\pi}\log(\frac{\left \| W_i \right \|_2 }{\left \| W_j \right \|_2})).
\nonumber
\end{split}
\end{equation}

Fortunately, our work addresses the challenge of quantifying class diversity by introducing the concept of information density. Here, we use information density to improve C2AM. Specifically, we modify $m_{ij}$ to:
\begin{equation}
\begin{split}
m_{ij}=\max(0,\frac{\alpha}{\pi}(\frac{I_i}{I_j})),
\nonumber
\end{split}
\end{equation}
where
\begin{equation}
\begin{split}
I_i = \frac{e^{I_i /(\bar{I}\cdot \sqrt{C} )}}{\sum_{j=1}^{C} e^{I_j / (\bar{I}\cdot \sqrt{C})}}\cdot C+1, \quad \bar{I} = \sum_{i=1}^{C} I_i.
\nonumber
\end{split}
\end{equation}
In this formula, the information density \( I_i \) is used to replace the weight norm ratios in the original method, thus more accurately reflecting the impact of class diversity on the decision space.

With this, we have detailed how to embed information density into loss functions for object detection. However, we still face an engineering challenge: the information density of classes changes with the model parameters, requiring dynamic updates. Calculating information density requires embeddings of all instances in a class, but we can only obtain a small number of samples within each batch during each iteration. If we were to re-extract embeddings of all instances in the dataset during every iteration, it would interrupt training and incur significant time costs. In the next section, we propose a novel training framework to address this issue.

\subsection{Low-Cost Dynamic Update of Information Density}
\label{sec5.2}

\subsubsection{Dynamic Update Strategy}
\label{sec5.2.1}

\textbf{The phenomenon of feature slow drift} \cite{ma2023delving} indicates that as training progresses, the distance between embeddings of the same sample at different times becomes smaller, making it possible to approximate the latest embedding with a previous version. Inspired by this, we propose a straightforward solution: store all instance embeddings generated during training in a queue, with the queue length equal to the total number of instances in the dataset. After each training epoch, update all embeddings in the queue. At the end of each epoch, use the embeddings in the queue to calculate and update the information density of all categories. We refer to this approach as the original strategy. While the original strategy avoids repeatedly extracting instance embeddings from the entire dataset, it increases storage space requirements.

Considering that calculating the class diversity in information density essentially requires the covariance matrix of instance embeddings, we propose a new strategy that significantly reduces storage space. The core idea of the new strategy is to use multiple local sample covariance matrices to calculate the global sample covariance matrix, which we prove in Proof 1. The specific steps are as follows:
\begin{itemize}[]
\item[(1)] Initialize a queue to store instance embeddings. The length of the queue can be adjusted according to the GPU memory size. Suppose the object detection dataset contains \( C \) categories with a total of \( N \) instances, and the queue length is \( d \). If \( d < N \), it means the queue cannot hold all instance embeddings. In this work, we set the queue length to 50,000, which can store 50,000 instance embeddings.

\item[(2)] At the beginning of a training epoch, first store the instance embeddings generated in each batch into the queue until it is full (i.e., storing \( d \) instance embeddings). Then, use the embeddings in the queue to calculate the local covariance matrix and mean for each category. Continuously update the queue, and once all old embeddings in the queue are updated, calculate the local covariance matrix and mean for each category again. By the end of an epoch, we can calculate \( \lfloor N/d \rfloor + 1 \) local sample covariance matrices \( \Sigma_i^k \) and means \( \mu_i^k \) for each category \( i = 1, \ldots, C \), \( k = 1, \ldots, \lfloor N/d \rfloor + 1 \).

\item[(3)] At the end of an epoch, use the stored local covariance matrices to calculate and update the information density for each category. Taking category \( i \) as an example, first calculate the global mean:
\begin{equation}
\begin{split}
\mu_i = \frac{1}{N_i} \sum_{k=1}^{\lfloor N/d \rfloor + 1} n_i^k \mu_i^k,
\nonumber
\end{split}
\end{equation}

where \( N_i \) is the total number of instances in category \( i \), and \( n_i^k \) is the number of instances in the local sample. Then, calculate the global covariance matrix:
\begin{small}
\begin{equation}
\begin{split}
\Sigma_i = \frac{1}{N_i} \left( \sum_{k=1}^{\lfloor N/d \rfloor + 1} n_i^k \Sigma_i^k + \sum_{k=1}^{\lfloor N/d \rfloor + 1} n_i^k (\mu_i^k - \mu_i)(\mu_i^k - \mu_i)^T \right). 
\nonumber
\end{split}
\end{equation}
\end{small}

The proof of this formula is provided in Proof~1. By integrating local covariance matrices to obtain the global covariance matrix, we significantly reduce the additional storage space required to update the information density. Further, the diversity of category \( i \) is estimated as \( \text{Vol}_i = \frac{1}{2} \log_2 \det(I + \Sigma_i) \), and the information density is \( I=Vol_i/A_i \), where \( A_i \) is the total area of instances in category \( i \).

\end{itemize}

\begin{algorithm}[t]
\setstretch{1} 
\caption{Dynamic Update of Information Density}
\label{alg:dynamic_info_density}
\begin{algorithmic}[1]
\STATE \textbf{Input:} Number of classes $C$, total number of instances $N$, queue length $d$ (e.g., 50,000), number of epochs $E$
\STATE Initialize queue $Q$ with maximum length $d$
\FOR{each epoch $e = 1$ to $E$}
    \FOR{each batch of data}
        \STATE Extract instance embeddings for each class and store in queue $Q$
        \IF{Queue $Q$ is full}
            \STATE Compute local covariance matrix $\Sigma_i^k$ and mean $\mu_i^k$ for each class $i$
            \STATE Update queue $Q$ with new embeddings
        \ENDIF
    \ENDFOR
    \STATE After processing all batches, compute additional local covariance matrices and means as needed
    \FOR{each class $i = 1$ to $C$}
        \STATE Compute global mean $\mu_i$:
        \[
        \mu_i \gets \frac{1}{N_i} \sum_{k=1}^{\lfloor N/d \rfloor + 1} n_i^k \mu_i^k
        \]
        \STATE Compute global covariance matrix $\Sigma_i$:
        \begin{align*}
        \Sigma_i \gets & \frac{1}{N_i} \bigg( \sum_{k=1}^{\lfloor N/d \rfloor + 1} n_i^k \Sigma_i^k \\
        & + \sum_{k=1}^{\lfloor N/d \rfloor + 1} n_i^k (\mu_i^k - \mu_i)(\mu_i^k - \mu_i)^T \bigg)
        \end{align*}
        \STATE Estimate the information density $I_i$ using Algorithm 1.
    \ENDFOR
\ENDFOR
\STATE \textbf{Output:} Information densities $I_i$ for each class $i$
\end{algorithmic}
\end{algorithm}

\begin{proof}1: Integrating Local Covariance Matrices to Obtain the Global Covariance Matrix.
Assume we have a dataset containing \( N \) instances, and we divide these instances into \( K \) batches, each containing \( n_k \) instances. For the \( k \)-th batch, let the instances be \(\{x_{k1}, x_{k2}, \ldots, x_{kn_k}\}\). The mean vector and local covariance matrix for this batch are defined as follows:
\begin{equation}
\mu_k = \frac{1}{n_k} \sum_{i=1}^{n_k} x_{ki},
\nonumber
\end{equation}

\begin{equation}
\Sigma_k = \frac{1}{n_k} \sum_{i=1}^{n_k} (x_{ki} - \mu_k)(x_{ki} - \mu_k)^T.
\nonumber
\end{equation}

The global covariance matrix is the covariance matrix of all batches, defined as:
\begin{equation}
\mu = \frac{1}{N} \sum_{k=1}^K \sum_{i=1}^{n_k} x_{ki},
\nonumber
\end{equation}

\begin{equation}
\Sigma = \frac{1}{N} \sum_{k=1}^K \sum_{i=1}^{n_k} (x_{ki} - \mu)(x_{ki} - \mu)^T.
\nonumber
\end{equation}

First, calculate the global mean \( \mu \):
\begin{equation}
\mu = \frac{1}{N} \sum_{k=1}^K \sum_{i=1}^{n_k} x_{ki} = \frac{1}{N} \sum_{k=1}^K n_k \mu_k.
\nonumber
\end{equation}

Then, split \((x_{ki} - \mu)\) in the global covariance matrix \( \Sigma \) into \((x_{ki} - \mu_k)\) and \((\mu_k - \mu)\):
\begin{equation}
\Sigma = \frac{1}{N} \sum_{k=1}^K \sum_{i=1}^{n_k} [(x_{ki} - \mu_k + \mu_k - \mu)(x_{ki} - \mu_k + \mu_k - \mu)^T].
\nonumber
\end{equation}
Expanding this, we get:
\begin{equation}
\begin{split}
\Sigma = \frac{1}{N} \sum_{k=1}^K &\sum_{i=1}^{n_k}[(x_{ki} - \mu_k)(x_{ki} - \mu_k)^T + (x_{ki} - \mu_k)(\mu_k - \mu)^T \\
& + (\mu_k - \mu)(x_{ki} - \mu_k)^T + (\mu_k - \mu)(\mu_k - \mu)^T].
\nonumber
\end{split}
\end{equation}

According to the properties of the covariance matrix, the first term is the local covariance matrix \( \Sigma_k \), and the expectation values of the second and third terms are zero. The fourth term can be calculated as:
\begin{equation}
\sum_{i=1}^{n_k} (\mu_k - \mu)(\mu_k - \mu)^T = n_k (\mu_k - \mu)(\mu_k - \mu)^T.
\nonumber
\end{equation}
Finally, the expression for the global covariance matrix is:
\begin{equation}
\begin{split}
\Sigma = \frac{1}{N} \left( \sum_{k=1}^K n_k \Sigma_k + \sum_{k=1}^K n_k (\mu_k - \mu)(\mu_k - \mu)^T \right).
\nonumber
\end{split}
\end{equation}

This formula demonstrates that the global covariance matrix can be calculated by taking a weighted sum of the local covariance matrices and adding the difference terms between local means and the global mean. This integration method effectively utilizes the unbiasedness and independence of the local covariance matrices, ensuring the accuracy of the global covariance matrix.
\label{proof1}
\end{proof}

\subsubsection{Storage Space Comparison}
\label{sec5.2.2}

Assume the object detection dataset contains \( N \) instances, each with an embedding dimension of \( p \), and there are \( C \) categories. The queue length is set to \( d \). The storage space required by the original strategy is:
\begin{equation}
\begin{split}
S_{\text{original}} = N \times p.
\nonumber
\end{split}
\end{equation}
The storage space required by the new strategy is:
\begin{equation}
\begin{split}
S_{\text{new}} = d \times p + C \times (\lfloor N/d \rfloor + 1) \times p^2.
\nonumber
\end{split}
\end{equation}
where \( C \times (\lfloor N/d \rfloor + 1) \times p^2 \) represents the space needed to store the local covariance matrices. To analyze when the new strategy saves more space, we define the storage space ratio \( R \):
\begin{equation}
\begin{split}
R = \frac{S_{\text{new}}}{S_{\text{original}}} = \frac{d \times p + C \times (\lfloor N/d \rfloor + 1) \times p}{N}.
\nonumber
\end{split}
\end{equation}

\begin{figure*}[t]
\centering
	\begin{minipage}{0.492\linewidth}
		\centering
		\includegraphics[width=1\linewidth]{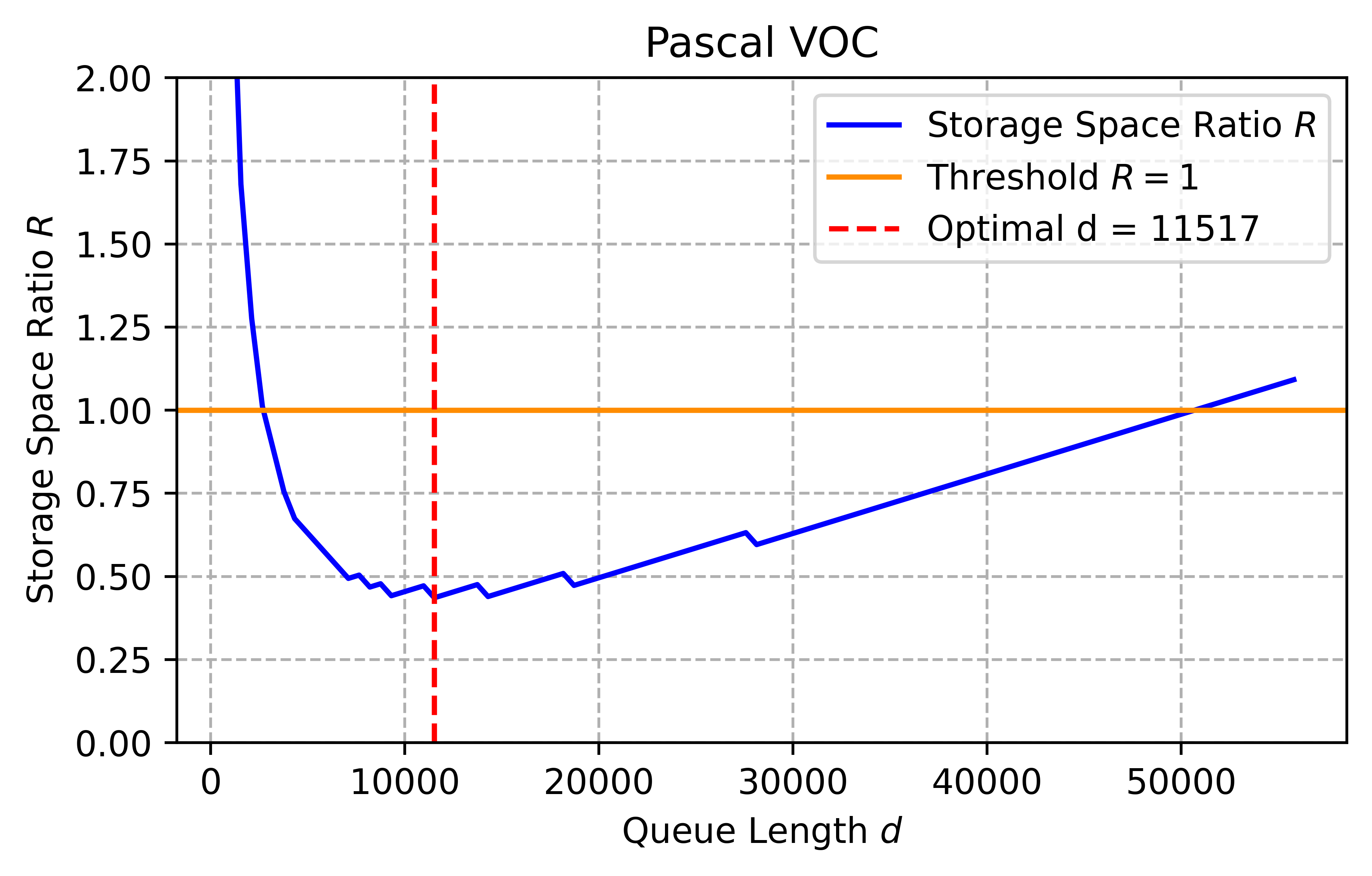}
	\end{minipage}
	\begin{minipage}{0.492\linewidth}
		\centering
		\includegraphics[width=1\linewidth]{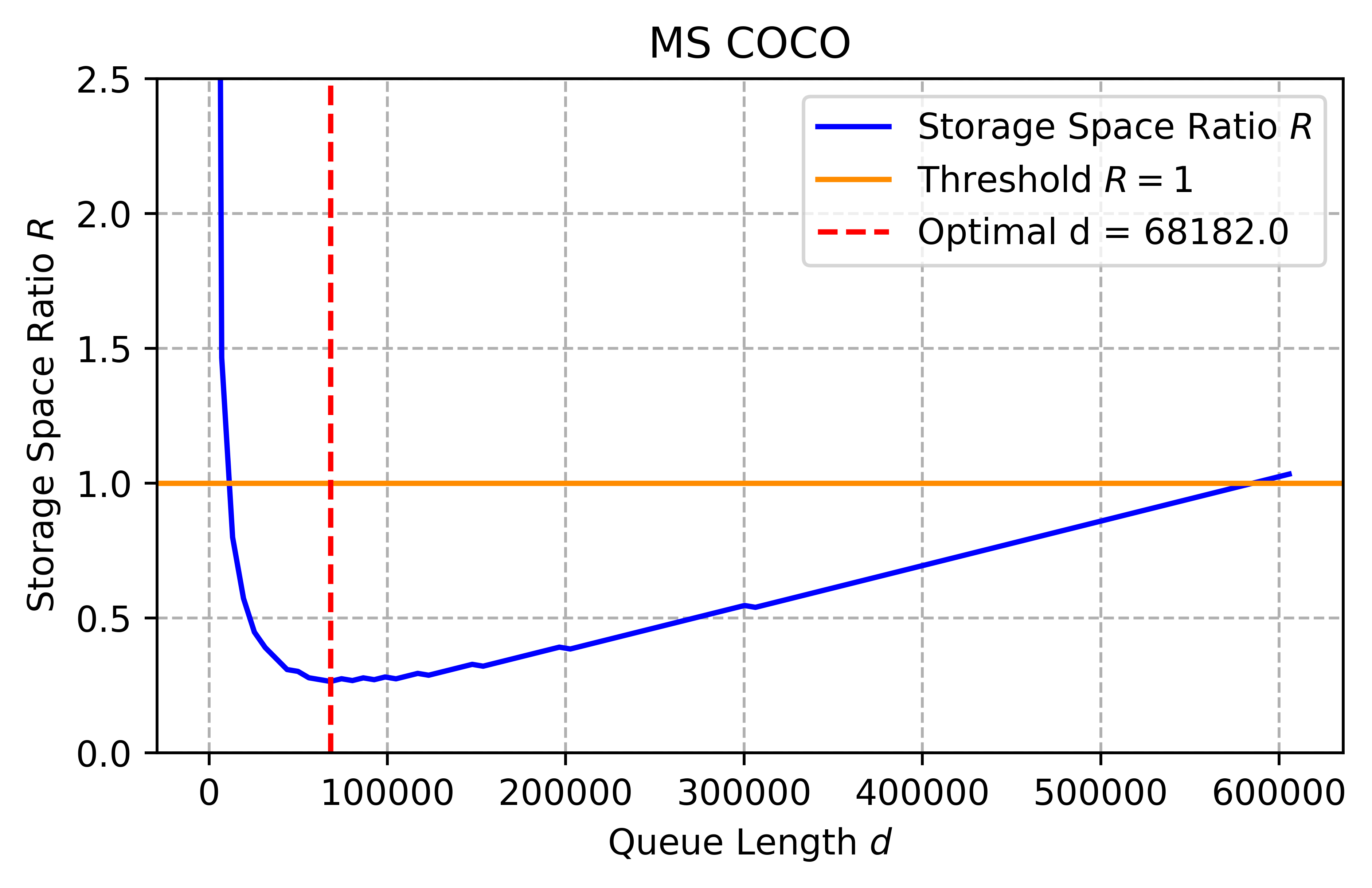}
	\end{minipage}
\caption{The function of the storage space ratio $R$ as it varies with the queue length $d$ on the Pascal VOC and MS COCO datasets.}
\label{fig5}
\end{figure*}

When \( R < 1 \), the new strategy saves more storage space. To visually compare the storage space requirements of the new and original strategies, we take the Pascal VOC, MS COCO, and LVIS v1.0 datasets as examples and plot the function graph of the storage space ratio \( R \) as it varies with the queue length \( d \) in Fig.~\ref{fig5}. It can be observed that in most cases, the storage space required by the new strategy is less than that of the original strategy. By visualizing our proposed storage space ratio, it becomes easy to choose the optimal queue length \( d \), thereby saving approximately 60\% of the storage space.
Given two examples \(\{N=55800, p=128, C=20\}\) and \(\{N=605638, p=128, C=80\}\), corresponding to the Pascal VOC and MS COCO datasets respectively, we use Listing \ref{lst:listing} to select the optimal queue lengths \( d \) of 11517 and 68182. For Example 1, the original strategy requires an additional 27.25 MB of memory, while the new strategy only requires 11.87 MB, saving approximately $\textbf{56.44\%}$ of memory. For Example 2, the new strategy can reduce memory usage from 295.72 MB to 78.29 MB, saving approximately $\textbf{73.52\%}$ of memory.

\begin{listing}[h]%
\caption{Selecting the optimal queue length $d$.}%
\label{lst:listing}%
\begin{lstlisting}[language=Python]
import matplotlib.pyplot as plt
import numpy as np

# Define parameters
N = 55800  # Total number of instances
p = 128     # Embedding dimension
C = 20      # Number of categories

# Define the range of queue lengths
d_values = np.linspace(1000, N, 100)
R_values = list((d_values + C * (np.floor(N / d_values) + 1) * p) / N)

min_R_values = min(np.abs(R_values))
optimal_d = d_values[R_values.index(min_R_values)]
print('The optimal d is:', optimal_d)
\end{lstlisting}
\end{listing}

The new training framework significantly reduces storage space utilization by merging the local sample covariance matrices, ensuring that the calculated value of information density remains unchanged. This innovative strategy not only provides an efficient solution for dynamically updating information density but also offers a low-cost storage solution for future research. In the experimental section, we demonstrate the significant improvement effects of information density on multiple loss functions. Additionally, loss functions improved with information density can be combined with other types of methods to further enhance the performance of object detection models (see Section \ref{sec6}).

\section{Experiments}
\label{sec6}

We conducted a comprehensive evaluation of the universal and significant improvements that information density brings to existing object detection loss functions on three object detection benchmark datasets. The experiments were divided into two parts. The first part was conducted on the non-long-tailed object detection dataset Pascal VOC. The second part was conducted on the long-tailed object detection datasets COCO-LT and LVIS v1.0.

\subsection{Datasets and Evaluation Metrics}
\label{sec6.1}

Pascal VOC \cite{VOC} includes two versions, 2007 and 2012, comprising a total of 20 classes. Following standard practice \cite{tong2023rethinking}, we trained on the train+val sets of VOC2007 and VOC2012 and tested on the test set of VOC2007. The COCO-LT \cite{number4} dataset is a long-tailed subset of MS COCO \cite{MSCOCO}, and they share the same validation set. Consistent with previous work \cite{seesaw}, we divided the 80 classes in COCO-LT into four groups based on the number of training instances per class: fewer than 20 images, 20$\sim$400 images, 400$\sim$8000 images, and 8000 or more images. We report the \emph{Average Precision (AP)} for each class on Pascal VOC, and on COCO-LT, we report the accuracy of the four groups as $AP_1^b$, $AP_2^b$, $AP_3^b$, and $AP_4^b$. The mean average precision is reported as $mAP^b$.
LVIS v1.0 \cite{lvis} contains 1,203 categories, with the training set consisting of 100k images (approximately 1.3M instances) and the validation set containing 19.8k images. Based on the frequency of occurrence in the training set, all categories are divided into three groups: rare (1$\sim$10 images), common (11$\sim$100 images), and frequent (more than 100 images). In line with EFL \cite{efl}, we report not only the widely used object detection metric $AP^b$ across IOU thresholds (from 0.5 to 0.95) but also the bounding box AP for frequent ($AP_f$), common ($AP_c$), and rare ($AP_r$) categories separately.

\begin{table}[t]
    \centering
    \caption{Average precision (\%) of Faster R-CNN with R-50-FPN backbone trained using various loss functions on the Pascal VOC.}
    \label{tab1}
\begin{small}
\renewcommand\arraystretch{1.2}
\setlength{\tabcolsep}{1.2pt} 
\begin{tabular}{lccccccc}
    \hline  \toprule
    \textbf{Class} & Seesaw & IDG-Seesaw & EFL & IDG-FL & C2AM & IDG-C2AM \\
    \midrule
    cat           & 89.5 & 89.0 & 88.3 & 88.7 & 88.1 & 87.8 \\
    car           & 79.8 & 82.5 & 78.9 & 80.5 & 80.8 & 81.3 \\
    horse         & 85.9 & 86.7 & 86.3 & 85.5 & 87.0 & 88.1 \\
    bus           & 84.8 & 85.8 & 82.9 & 81.9 & 84.5 & 85.2 \\
    bicycle       & 86.1 & 87.5 & 84.5 & 83.8 & 85.2 & 84.6 \\
    dog           & 88.4 & 88.2 & 85.3 & 86.0 & 86.5 & 87.4 \\
    person        & 79.1 & 79.7 & 77.1 & 80.1 & 77.8 & 79.5 \\
    train         & 84.2 & 85.9 & 82.9 & 84.4 & 85.3 & 87.0 \\
    motorbike     & 83.6 & 84.7 & 80.1 & 82.7 & 82.0 & 83.2 \\
    cow           & 82.9 & 83.5 & 79.8 & 81.6 & 80.7 & 82.1 \\
    \rowcolor{purple!10} 
    aeroplane     & 69.7 & 72.8 \textcolor{red}{$\uparrow$3.1} & 72.1 & 74.4 \textcolor{red}{$\uparrow$2.3} & 71.8 & 74.3 \textcolor{red}{$\uparrow$2.5} \\
    tvmonitor     & 75.8 & 79.3 & 76.4 & 77.2 & 74.1 & 75.7 \\
    sheep         & 76.5 & 78.7 & 74.8 & 76.9 & 75.6 & 76.5 \\
    bird          & 79.3 & 81.5 & 75.2 & 76.8 & 79.7 & 81.0 \\
    diningtable   & 74.0 & 76.0 & 71.3 & 73.2 & 73.4 & 75.1 \\
    sofa          & 78.7 & 79.1 & 72.7 & 70.5 & 79.5 & 80.8 \\
    \rowcolor{purple!10} 
    boat          & 66.4 & 68.2 \textcolor{red}{$\uparrow$1.8} & 62.4 & 64.9 \textcolor{red}{$\uparrow$2.5} & 64.2 & 66.5 \textcolor{red}{$\uparrow$2.2} \\
    \rowcolor{purple!10} 
    bottle        & 53.6 & 57.9 \textcolor{red}{$\uparrow$4.3} & 50.8 & 54.6 \textcolor{red}{$\uparrow$3.8} & 54.8 & 58.2 \textcolor{red}{$\uparrow$3.4} \\
    \rowcolor{purple!10} 
    chair         & 65.8 & 68.5 \textcolor{red}{$\uparrow$2.7} & 61.1 & 63.7 \textcolor{red}{$\uparrow$2.6} & 62.3 & 65.1 \textcolor{red}{$\uparrow$2.8} \\
    \rowcolor{purple!10}  
    pottedplant   & 52.7 & 57.3 \textcolor{red}{$\uparrow$4.6} & 49.1 & 53.2 \textcolor{red}{$\uparrow$4.1} & 51.5 & 54.8 \textcolor{red}{$\uparrow$3.3} \\ \midrule
\rowcolor{gray!20}  
    Total   & 76.9 & 78.6 \textcolor{red}{$\uparrow$1.7}  & 74.6 & 76.0 \textcolor{red}{$\uparrow$1.4} & 76.2 & 77.7 \textcolor{red}{$\uparrow$1.5} \\
    \bottomrule \hline
\end{tabular}
\end{small}
\end{table}

\begin{table}[t]
    \centering
    \caption{Average precision (\%) of Faster R-CNN with R-101-FPN backbone trained using various loss functions on the Pascal VOC.}
    \label{tab2}
\begin{small}
\renewcommand\arraystretch{1.2}
\setlength{\tabcolsep}{1.2pt} 
\begin{tabular}{lccccccc}
    \hline  \toprule
    \textbf{Class} & Seesaw & IDG-Seesaw & EFL & IDG-FL & C2AM & IDG-C2AM \\
    \midrule
    cat           & 91.4 & 91.8 & 91.2 & 90.5 & 89.8 & 88.0 \\
    car           & 82.2 & 83.0 & 81.1 & 82.4 & 82.0 & 83.3 \\
    horse         & 86.5 & 85.6 & 86.6 & 87.1 & 87.6 & 87.1 \\
    bus           & 83.6 & 84.9 & 84.4 & 85.7 & 85.3 & 86.5 \\
    bicycle       & 87.0 & 88.3 & 85.8 & 85.5 & 85.5 & 86.2 \\
    dog           & 90.8 & 90.6 & 90.2 & 90.3 & 87.1 & 86.7 \\
    person        & 84.9 & 84.6 & 82.8 & 84.6 & 78.2 & 79.8 \\
    train         & 86.4 & 88.0 & 85.4 & 85.9 & 85.5 & 86.1 \\
    motorbike     & 87.4 & 87.7 & 84.1 & 85.0 & 82.6 & 83.6 \\
    cow           & 81.8 & 82.0 & 80.1 & 81.9 & 81.0 & 81.9 \\
    \rowcolor{purple!10} 
    aeroplane     & 73.1 & 75.5 \textcolor{red}{$\uparrow$2.4} &70.0 & 72.6 \textcolor{red}{$\uparrow$2.6} & 72.5 & 74.7 \textcolor{red}{$\uparrow$2.2} \\
    tvmonitor     & 75.2 & 76.8 & 69.9 & 72.2 & 77.0 & 78.5 \\
    sheep         & 80.4 & 81.7 & 78.2 & 80.9 & 75.5 & 77.4 \\
    bird          & 74.1 & 76.4 & 72.3 & 74.1 & 80.4 & 81.2 \\
    diningtable   & 75.5 & 75.9 & 73.3 & 75.1 & 74.3 & 76.1 \\
    sofa          & 77.2 & 78.0 & 72.5 & 74.8 & 79.7 & 80.6 \\
    \rowcolor{purple!10}
    boat          & 63.3 & 65.3 \textcolor{red}{$\uparrow$2.0} & 64.2 & 67.2 \textcolor{red}{$\uparrow$3.0} & 65.2 & 67.9 \textcolor{red}{$\uparrow$2.7} \\
    \rowcolor{purple!10} 
    bottle        & 52.4 & 55.6 \textcolor{red}{$\uparrow$3.2} & 51.5 & 54.0 \textcolor{red}{$\uparrow$2.5} & 55.6 & 58.6 \textcolor{red}{$\uparrow$3.0} \\
    \rowcolor{purple!10} 
    chair         & 62.3 & 65.6 \textcolor{red}{$\uparrow$3.3} & 61.9 & 64.2 \textcolor{red}{$\uparrow$2.3} & 63.5 & 66.0 \textcolor{red}{$\uparrow$2.5} \\
    \rowcolor{purple!10} 
    pottedplant   & 53.9 & 57.8 \textcolor{red}{$\uparrow$3.9} & 50.0 & 53.5 \textcolor{red}{$\uparrow$3.5} & 52.1 & 55.8 \textcolor{red}{$\uparrow$3.7} \\ \midrule
\rowcolor{gray!20}  
    Total   & 77.5 & 78.7 \textcolor{red}{$\uparrow$1.2}  & 75.8 & 77.3 \textcolor{red}{$\uparrow$1.5} & 77.0 & 78.3 \textcolor{red}{$\uparrow$1.3}  \\
    \bottomrule \hline
\end{tabular}
\end{small}
\end{table}

\subsection{Implementation Details}
\label{sec6.2}

We implemented the Faster R-CNN \cite{faster} detector using the MMDetection \cite{mmdetection} toolbox and adopted ResNet-50 and ResNet-101 \cite{resnet} with an FPN \cite{size2} structure as the backbone networks. During training, we set the batch size to 16 and the initial learning rate to 0.02, consistent with Seesaw \cite{seesaw}, EFL \cite{efl}, and C2AM \cite{c2am}. We trained the model using an SGD optimizer with a momentum of 0.9 and a weight decay rate of 0.0001 for 24 epochs. The learning rate was reduced to 0.002 and 0.0002 at the end of the 16th and 22nd epochs, respectively. In all experiments, we applied random horizontal flipping and multi-scale jittering for data augmentation. We did not use any test-time augmentations.

\begin{table}[t]
\caption{Evaluation results on COCO-LT. The $mAP^b$, $AP_1^b$, $AP_2^b$, $AP_3^b$, and $AP_4^b$ (\%) for each method are reported, with red arrows indicating performance improvements.}
\label{tab3}
\centering  
\begin{small}
\renewcommand\arraystretch{1.45}
\setlength{\tabcolsep}{4.5pt} 
\begin{tabular}{l|l|cccc}
\hline \toprule 
 Dataset   &\multicolumn{5}{c}{COCO-LT}     \\ \hline
   &$mAP^b$   &$AP_1^b$   &$AP_2^b$   &$AP_3^b$ &$AP_4^b$ \\ \hline

\multicolumn{2}{l}{\textbf{R-50-FPN}} \\ \hline
Cross-Entropy (CE) & 24.5  &0 &14.6 &\underline{\textbf{29.6}}  &32.9 \\ \hline
Seesaw \cite{seesaw}   & 23.9  &3.0  &14.5  &28.4  &32.3 \\ 
\rowcolor{purple!10}
IDG-Seesaw    & 24.9 \textcolor{red}{$\uparrow$1.0}  &8.2 \textcolor{red}{$\uparrow$5.2}  &16.7   &28.6   &32.1  \\  \hline

EFL \cite{efl} & 25.0 &3.8 &16.3  &29.5  &32.5  \\ 
\rowcolor{purple!10}
IDG-FL     & 26.3 \textcolor{red}{$\uparrow$1.3}  &8.5 \textcolor{red}{$\uparrow$4.7}  &19.3  &29.7   &32.8   \\    \hline

C2AM \cite{c2am}     & 24.7  &2.8  &15.6  &29.4  &32.3  \\  
\rowcolor{purple!10}
IDG-C2AM    & 25.8 \textcolor{red}{$\uparrow$1.1}  &8.3 \textcolor{red}{$\uparrow$5.5}  &17.8  &29.8  &32.6 \\  \hline

\multicolumn{2}{l}{\textbf{R-101-FPN}} \\ \hline
Cross-Entropy (CE) & 26.0  &0 &16.4 &31.4 &34.2  \\  \hline

Seesaw \cite{seesaw}   & 24.9  &3.2  &14.5 &30.0 &33.4    \\ 
\rowcolor{purple!10}
IDG-Seesaw    & 26.1 \textcolor{red}{$\uparrow$1.2}  &9.0 \textcolor{red}{$\uparrow$5.8}  &16.8 &30.7 &33.5  \\  \hline

EFL \cite{efl} & 25.4  &3.6  &16.5 &30.2 &32.8   \\ 
\rowcolor{purple!10}
IDG-FL     & 26.5 \textcolor{red}{$\uparrow$1.1}  &9.1 \textcolor{red}{$\uparrow$5.5}  &19.4 &30.6 &32.0   \\  \hline

C2AM \cite{c2am}     & 25.1  &2.9  &15.6 &30.2 &32.7  \\  
\rowcolor{purple!10}
IDG-C2AM      & 26.3 \textcolor{red}{$\uparrow$1.2}  &8.5 \textcolor{red}{$\uparrow$5.6}  &18.1 &30.7 &32.5 \\  

\bottomrule \hline
\end{tabular}
\end{small}
\end{table}

\subsection{Evaluation on Pascal VOC}
\label{sec6.3}

We first trained baseline models using Seesaw, EFL, and C2AM losses. Then, we trained the proposed IDG-Seesaw, IDG-FL, and IDG-C2AM under the same settings. The experimental results on Pascal VOC are presented in Table \ref{tab1} and \ref{tab2}. It can be observed that even for Seesaw, EFL, and C2AM, which are designed to mitigate model bias, our improvements led to an overall increase in accuracy. Specifically, when the backbone is R-50-FPN, IDG-Seesaw, IDG-FL, and IDG-C2AM improve the overall performance by 1.7\%, 1.4\%, and 1.5\%, respectively, compared to the original methods. When using ResNet-101 as the backbone, the improved methods enhance the overall performance by 1.2\%, 1.5\%, and 1.3\%, respectively.

\textbf{More importantly for this work}, our methods significantly improve the performance of poorly performing categories. We selected five representative underperforming categories for observation: aeroplane, boat, bottle, chair, and potted plant. When the backbone is R-50-FPN, our improvements led to performance gains of \textbf{3.1\%}, \textbf{1.8\%}, \textbf{4.3\%}, \textbf{2.7\%}, and \textbf{4.6\%} for Seesaw loss in these five categories, respectively. When using R-101-FPN as the backbone, our improvements enhanced Seesaw loss performance by \textbf{2.4\%}, \textbf{2.0\%}, \textbf{3.2\%}, \textbf{3.3\%}, and \textbf{3.9\%} in these categories, respectively. Particularly, for the two worst-performing categories, bottle, and potted plant, our improvements were the most pronounced. The experimental results demonstrate that although Seesaw, EFL, and C2AM losses are designed to address model bias, our methods can further focus on and identify the underperforming categories. This further proves that the proposed information density can more accurately measure the learning difficulty of categories.

\begin{table}[t]
\caption{Evaluation results on LVIS v1.0. The $mAP^b$, $AP_r$, $AP_c$, and $AP_f$ (\%) for each method are reported, with red arrows indicating performance improvements.}
\label{tab4}
\centering  
\begin{small}
\renewcommand\arraystretch{1.35}
\setlength{\tabcolsep}{9pt} 
\begin{tabular}{l|l|ccc}
\hline \toprule 
 Dataset   &\multicolumn{4}{c}{LVIS v1.0}     \\ \hline
   &$mAP^b$   &$AP_r$   &$AP_c$   &$AP_f$  \\ \hline

\multicolumn{2}{l}{\textbf{R-50-FPN}} \\ \hline
SCE \cite{faster} & 19.3  &1.1 &16.1 &\underline{\textbf{30.9}}   \\ 
BCE \cite{faster} & 19.5  &1.6 &16.6 &30.6   \\ 
EQL \cite{eql} & 21.8  &3.6 &21.1 &30.5   \\ 
DropLoss \cite{droploss} & 21.8  &5.2 &21.8 &29.1   \\  \hline

Seesaw \cite{seesaw}   & 24.8  &14.8  &22.7 &31.6    \\ 
\rowcolor{purple!10}
IDG-Seesaw    & 25.9 \textcolor{red}{$\uparrow$1.1}  &17.9 \textcolor{red}{$\uparrow$3.1}  &23.5   &31.9     \\  \hline

EFL \cite{efl} & 26.0 &18.6 &24.5  &30.8    \\ 
\rowcolor{purple!10}
IDG-FL     & 27.1 \textcolor{red}{$\uparrow$1.1}  &22.1 \textcolor{red}{$\uparrow$3.5}   &25.4  &31.0      \\    \hline

C2AM \cite{c2am}     & 25.4  &15.6  &24.2  &30.9    \\  
\rowcolor{purple!10}
IDG-C2AM    & 26.6 \textcolor{red}{$\uparrow$1.2}  &19.0 \textcolor{red}{$\uparrow$3.4}  &25.1  &31.4   \\  \hline

\multicolumn{2}{l}{\textbf{R-101-FPN}} \\ \hline
SCE \cite{faster} & 20.9  &1.0 &18.2 &32.7   \\ 
BCE \cite{faster} & 21.4  &2.0 &19.3 &32.3   \\ 
EQL \cite{eql} & 23.4  &4.5 &22.9 & 32.3  \\ 
DropLoss \cite{droploss} & 23.5  &5.9 &23.9 &30.7   \\  \hline

Seesaw \cite{seesaw}   & 26.6  &14.9  &25.2 &33.3     \\ 
\rowcolor{purple!10}
IDG-Seesaw    & 28.0 \textcolor{red}{$\uparrow$1.4}  &19.2 \textcolor{red}{$\uparrow$4.3}  &26.5 &33.5   \\  \hline

EFL \cite{efl} & 26.3  &19.3  &23.2 &32.6   \\ 
\rowcolor{purple!10}
IDG-FL     & 27.3 \textcolor{red}{$\uparrow$1.0}   &22.5 \textcolor{red}{$\uparrow$3.2}  &24.7 &32.1    \\  \hline

C2AM \cite{c2am}     & 26.5  &18.1  &25.5 &31.2  \\  
\rowcolor{purple!10}
IDG-C2AM     & 27.8 \textcolor{red}{$\uparrow$1.3}  &21.0 \textcolor{red}{$\uparrow$2.9}  &26.8 &31.7  \\  

\bottomrule \hline
\end{tabular}
\end{small}
\end{table}

\subsection{Evaluation on COCO-LT and LVIS v1.0}
\label{sec6.4}

The comparison results on COCO-LT are summarized in Table \ref{tab3}. When using the R-50-FPN backbone, IDG-Seesaw, IDG-FL, and IDG-C2AM improved mAP by 1.0\%, 1.3\%, and 1.1\%, respectively, compared to Seesaw, EFL, and C2AM. \textbf{More importantly}, our improvements led to increases in $AP_1^b$ (bounding box detection precision for rare categories) by \textbf{5.2\%}, \textbf{4.7\%}, and \textbf{5.5\%}, respectively. When using the R-101-FPN backbone, our modifications further enhanced $AP_1^b$ for Seesaw, EFL, and C2AM by \textbf{5.8\%}, \textbf{5.5\%}, and \textbf{5.6\%}, respectively. The performance improvements for rare categories were even more pronounced compared to overall performance gains.

Table \ref{tab4} presents the experimental results on LVIS v1.0. Our approach consistently demonstrated superior performance gains for rare categories, once again highlighting our method's focus on addressing categories that are traditionally underperforming. By improving performance for these categories, we significantly reduced model bias. Specifically, when using the R-50-FPN backbone, IDG-Seesaw, IDG-FL, and IDG-C2AM achieved improvements in the $AP_r$ metric by \textbf{3.1\%}, \textbf{3.5\%}, and \textbf{3.4\%}, respectively, compared to the original methods. With the R-101-FPN backbone, our method increased the $AP_r$ metric by \textbf{4.3\%}, \textbf{3.2\%}, and \textbf{2.9\%}, respectively. These experimental results collectively indicate that information density can more accurately measure class difficulty, thereby promoting the model's focus on learning from the underperforming classes.

\begin{figure*}[t]
\centering
	\begin{minipage}{0.497\linewidth}
		\centering
		\includegraphics[width=1\linewidth]{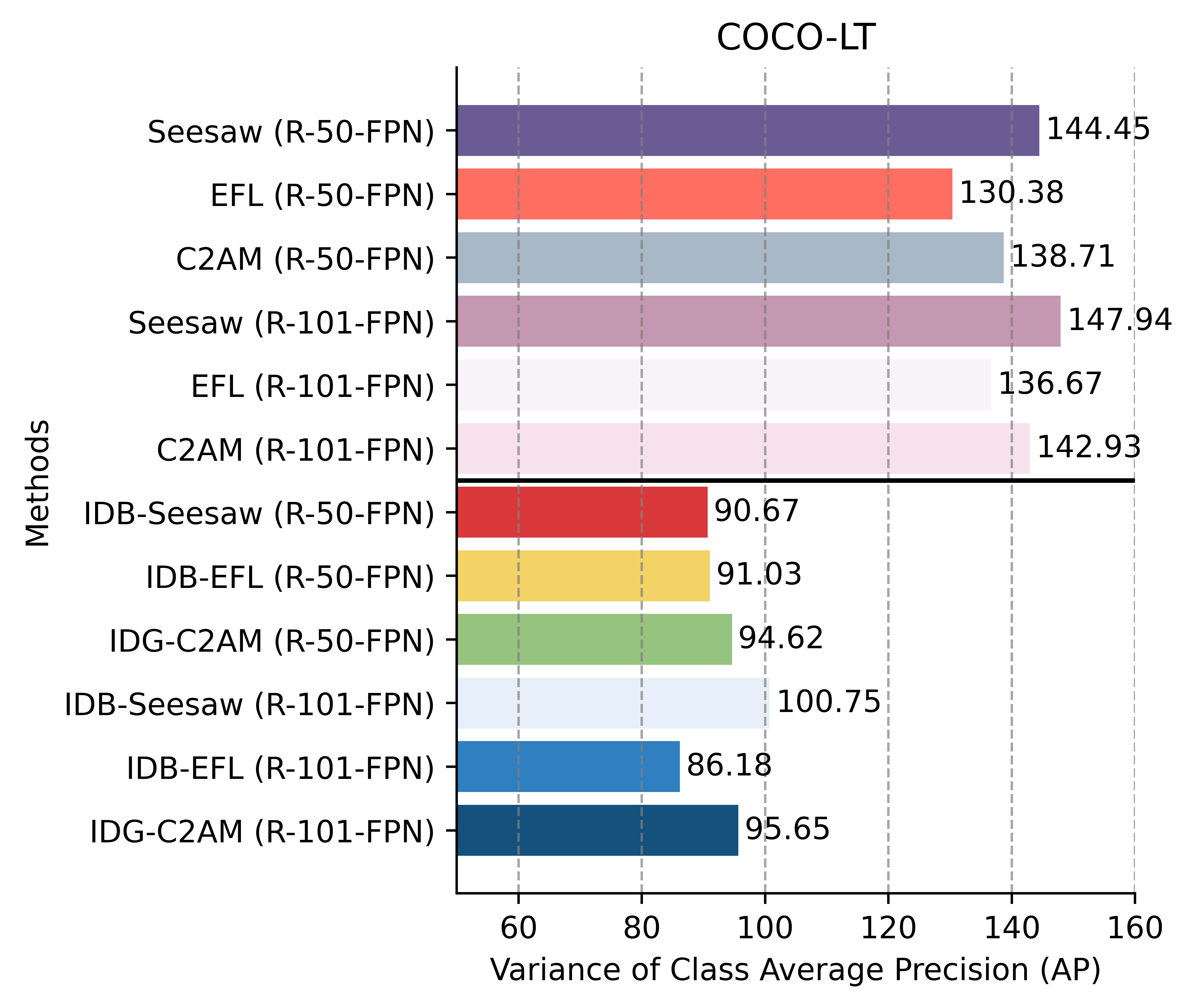}
	\end{minipage}
	\begin{minipage}{0.497\linewidth}
		\centering
		\includegraphics[width=1\linewidth]{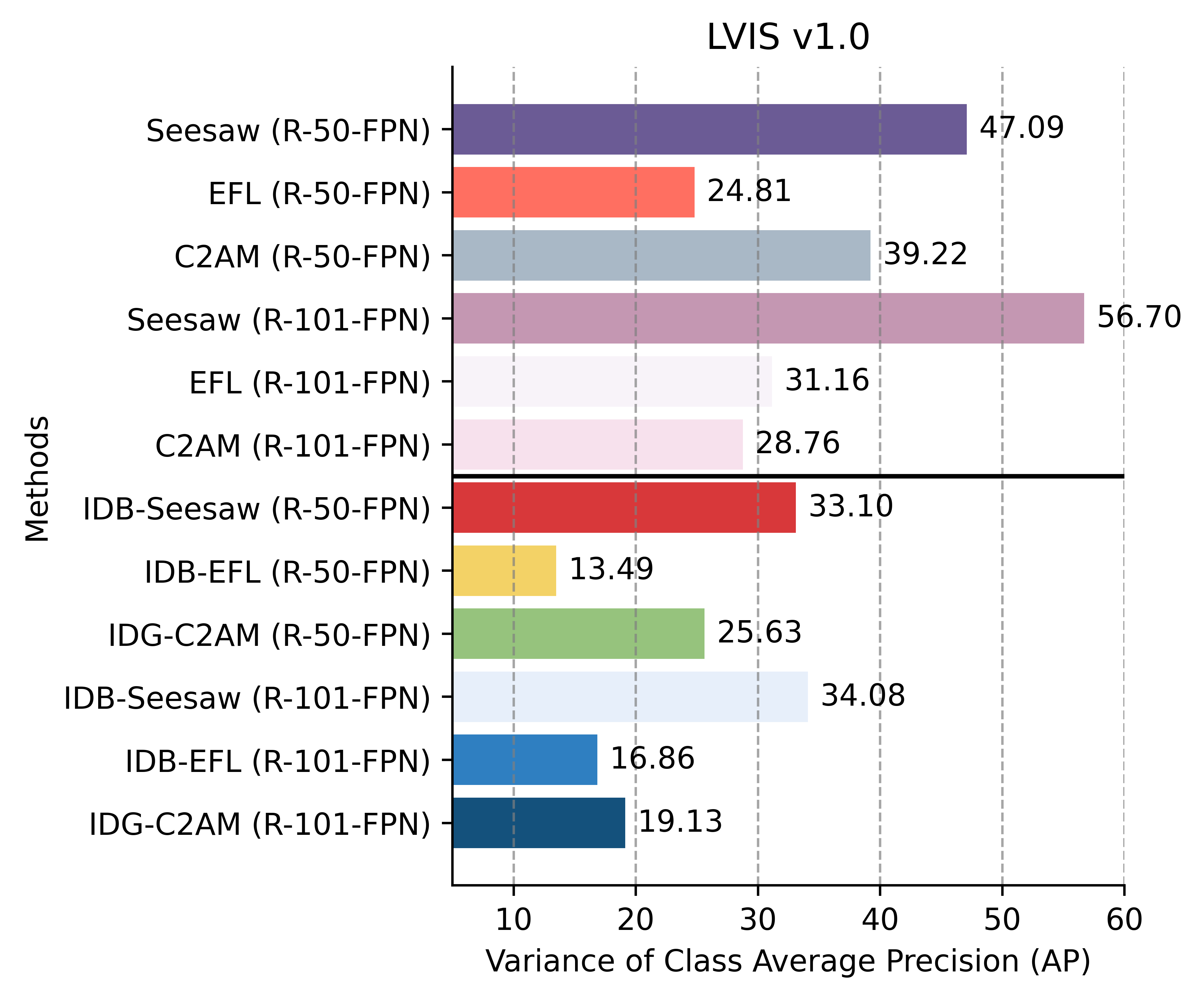}
	\end{minipage}
\caption{The effectiveness of our method in mitigating model bias on COCO-LT and LVIS v1.0.}
\label{fig6}
\end{figure*}

\subsection{Effectiveness in Reducing Model Bias}
\label{sec6.5}

To more clearly demonstrate the effectiveness of our method in mitigating model bias, we used the variance in Average Precision (AP) across classes as a measure of model bias. The comparative results on COCO-LT and LVIS v1.0 are presented in Fig.~\ref{fig6}. It can be observed that under two different backbones, our method significantly reduces model bias. For example, on COCO-LT, IDG-Seesaw (R-50-FPN) reduces model bias by \textbf{37.23\%} compared to Seesaw (R-50-FPN). On LVIS v1.0, IDG-FL (R-101-FPN) reduces model bias by \textbf{45.89\%} compared to EFL (R-101-FPN).

The significant improvements to Seesaw, EFL, and C2AM suggest that mitigating model bias by addressing information density imbalance is a parallel approach to addressing gradient imbalance or decision space imbalance. That is, alleviating gradient imbalance does not necessarily address model bias caused by information density imbalance. Therefore, we urge other researchers to consider the model bias introduced by information density imbalance when designing object detection models.

\section{Conclusion and Future Work}
\label{sec7}

This work investigates the issue of generalized bias in deep learning models for object detection tasks, which cannot be explained solely by the number of instances. We proposed the concept of information density to measure the difficulty of detecting different categories. Experiments revealed a significant negative correlation between a category’s information density and its accuracy. Subsequently, we utilized information density to improve three advanced object detection loss functions. The experimental results demonstrate that our improvements are most pronounced for categories with poor performance. Comprehensive empirical research indicates that information density provides a more accurate evaluation of a category’s learning difficulty, aiding models in focusing on learning underrepresented classes. In the future, we believe that information density could advance object detection tasks in the following directions:
\begin{itemize}[]
\item[(1)] Information density can guide data augmentation methods for object detection datasets. For categories with high information density, traditional data augmentation techniques like Copy-Paste or data expansion using generative models can be employed.

\item[(2)] Model Capability Exploration: Information density can be used to explore the capability limits of different object detection models or backbones. Higher information density implies that the model must handle larger amounts of information per unit, which may require larger models to tackle the task effectively. Quantifying the capability limits of each model will help engineers choose models with appropriate parameter sizes for different scale tasks, balancing deployment costs and performance requirements.

\item[(3)] Understanding and Improving Model Architecture: The high correlation between a category’s information density and its accuracy provides new insights into understanding the learning mechanisms of object detection models and improving model architectures. From the perspective of information compression, the ``capacity" of a model should be related to its performance. Therefore, when designing model architectures, additional adaptive modules or branches can be added to further refine and handle categories with high information density, indirectly increasing the model's ``capacity" for specific categories.
\end{itemize}

Additionally, aside from information density, we believe there are other underlying factors affecting model bias or performance. We hope to draw attention from other researchers to further enhance the interpretability of fairness in object detection models.

\bibliographystyle{IEEEtran}
\bibliography{egbib}



%




\end{document}